%% file: main.tex
\documentclass{ieeetj}
\usepackage{cite}
\usepackage{amsmath,amssymb,amsfonts}
\usepackage{algorithmic}
\usepackage{graphicx,color}
\usepackage{textcomp}
\usepackage{xcolor}
\usepackage{hyperref}
\hypersetup{hidelinks}
\usepackage{algorithm,algorithmic}
\def\BibTeX{{\rm B\kern-.05em{\sc i\kern-.025em b}\kern-.08em
    T\kern-.1667em\lower.7ex\hbox{E}\kern-.125emX}}
\AtBeginDocument{\definecolor{tmlcncolor}{cmyk}{0.93,0.59,0.15,0.02}\definecolor{NavyBlue}{RGB}{0,86,125}}

\let\labelindent\relax
\usepackage{enumitem}
\usepackage{booktabs}
\usepackage{subcaption}
\usepackage{float}
\usepackage{stfloats}

\usepackage[normalem]{ulem}
\newif\ifannotated

\annotatedtrue
\annotatedfalse

\ifannotated
\newcommand{\add}[1]{{\color{blue}{#1}}}
\newcommand{\delete}[1]{{\color{red}{\sout{#1}}}}

\newcommand{\margincomment}[1]{\marginpar{$\Rightarrow$\color{red}\fbox{\parbox{\linewidth}{\color{black}\scriptsize#1}}}}
\else
\newcommand{\add}[1]{{{#1}}}
\newcommand{\delete}[1]{\ignorespaces}

\newcommand{\margincomment}[1]{}
\fi

\def\OJlogo{\vspace{-4pt}$<$Society logo(s) and publication title will appear here.$>$}
\def\seclogo{\vspace{10pt}$<$Society logo(s) and publication title will appear here.$>$}

\def\authorrefmark#1{\ensuremath{^{\textbf{#1}}}}

\begin{document}
\receiveddate{XX Month, XXXX}
\reviseddate{XX Month, XXXX}
\accepteddate{XX Month, XXXX}
\publisheddate{XX Month, XXXX}
\currentdate{XX Month, XXXX}
\doiinfo{XXXX.2022.1234567}

\markboth{}{Author {et al.}}

\title{Parameter-efficient Multi-Task and Multi-Domain Learning using Factorized Tensor Networks}

\author{Yash Garg\authorrefmark{1}, Nebiyou Yismaw\authorrefmark{1}, Rakib Hyder\authorrefmark{1},\\ Ashley Prater-Bennette\authorrefmark{2}, Amit Roy-Chowdhury\authorrefmark{1}, and M. Salman Asif\authorrefmark{1}}
\affil{ University of California Riverside, CA 92508, USA}
\affil{Air Force Research Laboratory, Rome, NY 13441, USA}

\corresp{Corresponding author: M. Salman Asif (email: sasif@ucr.edu).}

\authornote{This work is supported in part by NSF CAREER award CCF-2046293, AFOSR award FA9550-21-1-0330, ONR award N00014-19-1-2264, and USDA award 2023-67021-40629.}

\begin{abstract}
Multi-task and multi-domain learning methods seek to learn multiple tasks/domains, jointly or one after another, using a single unified network. The primary challenge and opportunity lie in leveraging shared information across these tasks and domains to enhance the efficiency of the unified network. The efficiency can be in terms of accuracy, storage cost, computation, or sample complexity. In this paper, we introduce a factorized tensor network (FTN) designed to achieve accuracy comparable to that of independent single-task or single-domain networks, while introducing a minimal number of additional parameters. The FTN approach entails incorporating task- or domain-specific low-rank tensor factors into a shared frozen network derived from a source model. This strategy allows for adaptation to numerous target domains and tasks without encountering catastrophic forgetting.  Furthermore, FTN requires a significantly smaller number of task-specific parameters compared to existing methods. We performed experiments on widely used multi-domain and multi-task datasets. We show the experiments on convolutional-based architecture with different backbones and on transformer-based architecture. Our findings indicate that FTN attains similar accuracy as single-task or single-domain methods while using only a fraction of additional parameters per task. The code is available at \url{https://doi.org/10.24433/CO.7519211.v2}.
\end{abstract}

\begin{IEEEkeywords}
\vspace{-2ex}
Low-rank Adaptation, Multi-domain/Multi-task learning, Tensor Decomposition
\end{IEEEkeywords}


\maketitle

\input{sections/introduction}
\input{sections/related_work}
\input{sections/methods}
\input{sections/experiment}

\input{sections/conclusion}


{\small
\bibliographystyle{IEEEtran} 
\bibliography{egbib}
}

\clearpage

\input{supplementary}

\end{document}

%% file: sections/introduction.tex
\section{Introduction}

\begin{figure*}[t]
  \centering
  \includegraphics[width=\textwidth]{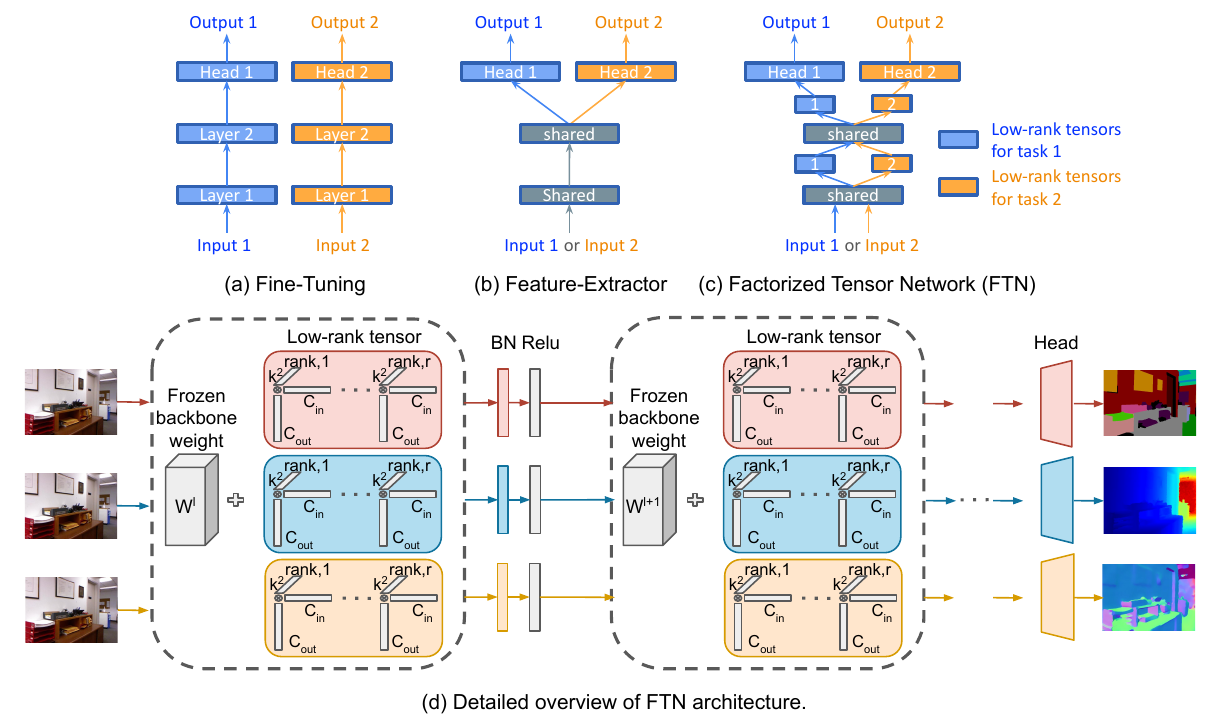}
  \caption{Overview of different MTL/MDL approaches and our proposed method. (a) Fine-Tuning trains entire network per task/domain. (b) Feature-Extractor trains a backbone network shared by all tasks/domains with task/domain-specific heads. (c) Our proposed method, Factorized Tensor Network (FTN), adapts to a new task/domain by adding low-rank factors to shared layers. (d) Detailed overview of FTN. A single network adapted to three downstream vision tasks (segmentation, depth, and surface normal estimation) by adding task-specific low-rank tensors ($\Delta \mathcal{W}_t$). Task/domain-specific blocks are shown in same colors.}
  \label{fig:overview_approach}
\end{figure*}

%
%
The primary objective in multi-task learning (MTL) is to train a single model that learns multiple related tasks, either jointly or sequentially. Multi-domain learning (MDL) aims to achieve the same learning objective across multiple domains. MTL and MDL techniques seek to improve overall performance by leveraging shared information across multiple tasks and domains. On the other hand, single-task or single-domain learning does not have that opportunity. Likewise, the storage and computational cost associated with single-task/domain models quickly grows as the number of tasks/domains increases. In contrast, MTL and MDL methods can use the same network resources for multiple tasks/domains, which keeps the overall computational and storage cost small \cite{wallingford2022task, rebuffi2018efficient}.

%
%
In general, MTL and MDL can have different input/output configurations, but we model them as task/domain-specific network representation problems. 
Let us represent a network for MTL or MDL as the following general function:
\begin{equation}\label{eq:MTL_MDL}
    \mathbf{y}_t = \mathbf{F}_t(\mathbf{x}) \equiv \mathbf{F}(\mathbf{x}; \mathcal{W}_t, h_t), 
\end{equation}
where $\mathbf{F}_t$ represents a function for task/domain $t$ that maps input $\mathbf{x}$ to output $\mathbf{y}_t$. We further assume that $\mathbf{F}$ represents a network with a fixed architecture and $\mathcal{W}_t$ and $h_t$ represent the parameters for task/domain-specific feature extraction and classification/inference heads, respectively. The function in \eqref{eq:MTL_MDL} can represent the network for specific task/domain $t$ using the respective $\mathcal{W}_t,h_t$. In the case of MTL, with $T$ tasks, we can have $T$ outputs $\mathbf{y}_1,\ldots, \mathbf{y}_T$ for a given input $\mathbf{x}$. In the case of MDL, we usually have a single output for a given input, conditioned on the domain $t$. 
Our goal is to learn the $\mathcal{W}_t,h_t$ for all $t$ that maximize the performance of MTL/MDL with minimal computation and memory overhead compared to single-task/domain learning. 

Figure~\ref{fig:overview_approach}(a),(b),(c) illustrate three typical approaches for MTL/MDL. 
First, we can start with a pre-trained network and fine-tune all the parameters $(\mathcal{W}_t)$ to learn a target task/domain, as shown in Figure \ref{fig:overview_approach}(a). Fine-Tuning approaches can transfer some knowledge from the pretrained network to the target task/domain, but they effectively use an independent network for every task/domain \cite{mallya2018piggyback, wallingford2022task}. 
Second, we can reduce the parameter and computation complexity by using a completely shared Feature-Extractor (i.e., $\mathcal{W}_t = \mathcal{W}_\text{shared}$ for all $t$) and learning task/domain-specific heads as last layers, as shown in Figure~\ref{fig:overview_approach}(b). While such approaches reduce the number of parameters, they often result in poor overall performance because of limited network capacity and interference among features for different tasks/domains \cite{mallya2018piggyback,  wallingford2022task, zhang2020side}. 
%
%
Third, we can divide the network into shared and task/domain-specific parameters or pathways, as shown in Figure~\ref{fig:overview_approach}(c). Such an approach can increase the network capacity, provide interference-free paths for task/domain-specific feature extraction, and enable knowledge sharing across the tasks/domains. In recent years, a number of such methods have been proposed for MTL/MDL  
\cite{wallingford2022task, kanakis2020reparameterizing, misra2016cross}. 
%
%
While existing methods can provide performance comparable to single-task/domain learning, they require a significantly large number of additional parameters.

In this paper, we propose a parameter-efficient approach to factorize a network into two distinct modules: a shared frozen module and a task/domain-specific module. We refer to this architecture as a factorized tensor network (FTN). FTNs adapt a network to target domains or tasks by learning low-rank tensors and normalization layers, such as batch normalization. An illustration of our proposed method is shown in Figure~\ref{fig:overview_approach}(d), where we represent network parameters as $\mathcal{W}_t = \mathcal{W}_\text{shared} + \Delta \mathcal{W}_t$, where $\Delta \mathcal{W}_t$ is a low-rank tensor. 

%

Similar parameter-efficient methods such as \cite{hu2021lora, he2023parameter}, use low-rank matrix adaptations to fine-tune their network. Our proposed method represents low-rank adaptations as a summation of R rank-1 tensors, significantly reducing the number of parameters in our network while achieving better performance. 
LoRA \cite{hu2021lora} explores a similar approach by using low-rank matrix factorization to adapt networks. However, unlike LoRA, the low-rank tensor factorization in FTN enables greater parameter reduction. Our experiments demonstrate that FTN achieves better results than LoRA. While LoRA was originally designed for transformer architectures, we have shown a natural extension of FTN to convolutional architectures.
The recent method SVFT \cite{lingam2024svft} updates weights as a sparse combination of outer products of singular vectors, training only the coefficients of these sparse combinations. Our experiments indicate that FTN achieves superior performance using fewer parameters than SVFT.
FTN leverages tensor factorization to efficiently approximate multi-dimensional data. \add{Our main motivation is to exploit the ability of tensor factorization to model complex interactions and dependencies more effectively than traditional 2D matrix representations \cite{rabanser2017introduction, kolda2009tensor}.}
\delete{while preserving structural relationships across different modes. This expressiveness allows FTN to model complex interactions and dependencies more effectively than traditional 2D matrix representations.}
%

A prior work, TAPS \cite{wallingford2022task}, differentially learns which layers of a pre-trained network to adapt for a downstream task/domain by learning an indicator function. The network uses adapted weights instead of pre-trained weights if the indicator score is above a certain threshold. This typically involves adapting high-parameterized layers closer to the classifier/head, which uses significantly more parameters than our FTN method. 
%
Existing parameter-efficient MTL/MDL methods \cite{mallya2018piggyback, mallya2018packnet, li2016revisiting}  introduce small task/domain-specific parameters while others \cite{zhang2020side, guo2019spottune} add many parameters to boost the performance irrespective of the task complexity. 
In our work, we demonstrate the flexibility of FTNs by selecting the rank according to the complexity of the task.
Other approaches like RCM \cite{kanakis2020reparameterizing} adapt incrementally to new tasks by reparameterizing the convolutional layer into task-shared and task-specific parameters. However, unlike FTN this architecture shows limitations in adapting based on the complexity of the tasks and performs subpar along performance and parameters.
We demonstrate the effectiveness of our method using different MTL and MDL datasets. 
\paragraph*{Contributions.}  
The main contributions of this paper are as follows. 
%
\begin{itemize}[leftmargin=*,noitemsep,topsep=0pt]
    \item We propose a new method for MTL and MDL, called factorized tensor networks (FTN), that adds task/domain-specific low-rank tensors to shared weights. FTNs can achieve similar performance as the single-task/domain methods while using a fraction of additional parameters. 
    \item Our proposed method utilizes tensor-factorization and demonstrates superior parameter-efficiency compared to matrix factorization methods such as LoRA \cite{hu2021lora} or indicator based adaptation methods such as TAPS \cite{wallingford2022task}. 
    \item Our proposed FTNs can be viewed as a plug-in module that can be added to any pretrained network and layer. We have shown this by extending FTNs to transformer-based architectures.
    \item We performed empirical analysis to show that the FTNs enable flexibility by allowing us to vary the rank of the task-specific tensors based on the problem complexity. 
    %
\end{itemize}

%% file: sections/related_work.tex
\section{Related Work}

\textbf{Multi-task learning (MTL)} methods commonly leverage shared and task-specific layers in a unified network to solve related tasks \cite{zhang2022rethinking, zhang2022automtl}. These methods learn shared and task-specific representation through their respective modules. Optimization-based methods \cite{chen2018gradnorm, chen2020just} devise a principled way to evaluate gradients and losses in multi-task settings. Branched and tree-structured MTL methods \cite{zhang2022tree} enable different tasks to share branches along a tree structure for several layers. Multiple tasks can share computations and features in any layer only if they belong to the same branch in all the preceding layers. \cite{kanakis2020reparameterizing, maninis2019attentive} proposed MTL networks that incrementally learn new tasks. ASTMT \cite{maninis2019attentive} proposed a network that emphasizes or suppresses features depending on the task at hand. RCM \cite{kanakis2020reparameterizing} reparameterizes the convolutional layer into non-trainable and task-specific trainable modules. We compare our proposed method with these incrementally learned networks. 
Adashare \cite{sun2020adashare} is another related work in MTL that jointly learns multiple tasks. It learns task-specific policies and network pathways \cite{jangcategorical}.


\noindent
\textbf{Multi-domain learning (MDL)} focuses on adapting one network to multiple unseen domains or tasks. MDL setup trains models on task-specific modules built upon the frozen backbone network. This setup helps MDL networks avoid negative transfer learning or catastrophic forgetting, which is common among multi-task learning methods. 
The work by \cite{rebuffi2017learning, rebuffi2018efficient} introduces the task-specific parameters called residual adapters. The architecture introduces these adapters as a series or parallel connection on the backbone for a downstream task. Inspired by pruning techniques, Packnet \cite{mallya2018packnet} learns on multiple domains sequentially on a single task to decrease the overhead storage, which comes at the cost of performance. Similarly, the Piggyback \cite{mallya2018piggyback} method uses binary masks as the module for task-specific parameters. These masks are applied to the weights of the backbone to adapt them to new domains. To extend this work, WTPB \cite{mancini2018adding} uses the affine transformations of the binary mask on their backbone to extend the flexibility for better learning. BA$^{2}$ \cite{berriel2019budget} proposed a budget-constrained MDL network that selects the feature channels in the convolutional layer. It gives a parameter-efficient network by dropping the feature channels based on budget but at the cost of performance. \add{DA3~\cite{yang2022da3} introduces a memory- and parameter-efficient method with a specific focus on on-device applications. DA3 freezes multiplicative weights and masks and only updates the additive bias terms.} \cite{zhao2021and} paper learns the adapter modules and the plug-in architecture of the modules using NAS. Spot-Tune \cite{guo2019spottune} learns a policy network, which decides whether to pass each image through Fine-Tuning or pretrained networks. It neglects the parameter efficiency factor and emphasises more on performance. TAPS \cite{wallingford2022task} adaptively learns to change a small number of layers in a pretrained network for the downstream task. 

\noindent \textbf{Domain adaptation and transfer learning.} The work in this field usually focuses on learning a network from a given source domain to a closely related target domain. The target domains under this kind of learning typically have the same category of classes as source domains \cite{tzeng2017adversarial}. Due to this, it benefits from exploiting the labels of source domains to learn about multiple related target domains\cite{venkateswara2017deep}. Some work has a slight domain shift between source and target data, like different camera views \cite{saenko2010adapting}. At the same time, recent papers have worked on significant domain shifts like converting targets into sketch or art domains \cite{venkateswara2017deep, zhao2017stretching}. 
%
Transfer learning is related to MDL or domain adaptation but focuses on better generalizing target tasks \cite{mustafa2021supervised}. Most of the work in this field uses the popular ImageNet as a source dataset to learn feature representation and learn to transfer to target datasets. The method proposed in \cite{yang2022factorizing} uses a pretrained (multi-task) teacher network and decomposes it into multiple task/knowledge-specific factor networks that are disentangled from one another. This factorization leads to sub-networks that can be fine-tuned to downstream tasks, but they rely on knowledge transfer from a teacher network that is pretrained for multiple tasks. Modular deep learning methods \cite{pfeiffer2023modular} focus on transfer learning by avoiding negative task interference and having parameter-efficient modules.  

\noindent
\textbf{Factorization methods in MDL/MTL.} The method in \cite{yang2015unified} proposed a unified framework for MTL/MDL using semantic descriptors, without focusing on parameter-efficient adaptation. \cite{yang2017deep} performs MTL/MDL by factorizing each layer in the network after incorporating task-specific information along a separate dimension. Both the networks in \cite{yang2015unified} and \cite{yang2017deep} require retraining from scratch for new tasks/domains. In contrast, FTN can incrementally learn low-rank factors to add new tasks/domains. \cite{chen2018sharing} proposed a new parameter-efficient network to replace residual networks by incorporating factorized tensors. The results in \cite{chen2018sharing} are limited to learning single-task networks, where the network is only compressed by up to $\sim 60\%$. In \cite{bulat2020incremental}, the authors proposed a network for MDL using Tucker decomposition. \add{\cite{yismaw2024domain} paper focuses on solving inverse problems in computational imaging applications. The method proposes to modulate the weights of an unrolled pre-trained network
for adaptation to multiple domains, measurement models, and noise. The multiplicative modulation is applied on DCNN (a small parameter network) with only rank-1 tensors.}

\noindent
\textbf{Transformer-based methods in MDL/MTL.} \add{COMPACTER~\cite{karimi2021compacter} is a parameter-efficient fine-tuning method designed for large-scale language models. It inserts task-specific weight matrices into a pretrained model’s weights as a sum of Kronecker products between shared low-rank ”slow” weights and task-specific ”fast” rank-one matrices. Adaptformer~\cite{chen2022adaptformer} introduces an effective adapter-based approach for parameter-efficient fine-tuning of vision transformers for a large variety of downstream visual recognition tasks. The core idea is to insert the lightweight bottleneck adapters into the feed-forward layer of a pretrained transformer. The adapter involves two fully connected layers, a non-linear activation function, and a scaling factor.} LoRA \cite{hu2021lora} is a low-rank adaptation method proposed for large language models, which freezes the pre-trained weights of the model and learns low-rank updates for each transformer layer. It updates weight matrices for query and value in every attention layer. Similarly, KAdaptation \cite{he2023parameter} proposes a parameter-efficient adaptation method for vision transformers. It represents the updates of MHSA layers using the summation of Kronecker products between shared parameters and low-rank task-specific parameters. We compared both of these methods and have shown that FTN outperforms along the number of parameters. Scaling and shifting your features (SSF) \cite{lian2022scaling} is another transformer method for parameter-efficient adaptation that applies element-wise multiplication and addition to tokens after different operations. SSF, in principle, is similar to fine-tuning the Batch Normalization layer in convolutional layers, which has scaling and shifting trainable parameters. FTN trains the Batch Normalization layers and has the same effect as scaling and shifting features when adapting to new tasks. \cite{ye2022inverted} proposed inverted-pyramid multi-task transformer, performs cross-task interaction among spatial features of different tasks in a global context. 
Our method, FTN, shares some high-level similarities with other parameter-efficient adaptation methods such as LoRA, as both approaches aim to introduce low-rank factors to adapt networks for multiple tasks and domains. Our method is a natural extension to higher-order tensors, and we demonstrate its effectiveness across both transformer and convolutional network architectures. In addition, our method adds parameter and performance efficiency compared to related methods, as shown by our experiments.


In summary, our proposed method (FTN) offers a parameter-efficient approach to achieve performance comparable to or better than existing adaptation methods by utilizing a fraction of additional parameters. Our primary design consideration was to achieve efficient adaptation, enabling incremental learning with additive factors.
To achieve parameter efficiency, we introduce a small number of trainable parameters through low-rank factorization applicable to both convolutional and transformer-based networks. We utilize frozen and trainable task-specific parameters to support incremental learning without forgetting prior knowledge.

%% file: sections/methods.tex
\section{Technical Details}

\noindent\textbf{Notations.} 
In this paper, we denote scalars, vectors, matrices and tensors by $w$, $\mathbf{w}$, $\mathrm{W}$, and $\mathbf{W}$, respectively. The collective set of tensors (network weights) is denoted as $\mathcal{W}$.

\subsection{FTN applied to Convolutional layers}



In our proposed method, we use task/domain-specific low-rank tensors to adapt every convolutional layer of a pretrained backbone network to new tasks and domains. Let us assume the backbone network has $L$ convolutional layers that are shared across all task/domains. We represent the shared network weights as $\mathcal{W}_\text{shared} = \{\mathbf{W}_1, \ldots, \mathbf{W}_L\}$ and the low-rank network updates for task/domain $t$ as $\Delta \mathcal{W}_t = \{\Delta\mathbf{W}_{1,t}, \ldots, \Delta\mathbf{W}_{L,t}\}$. To compute features for task/domain $t$, we update weights at every layer as $\mathcal{W}_\text{shared}+\Delta \mathcal{W}_t = \{\mathbf{W}_1 + \Delta \mathbf{W}_{1,t},\ldots,\mathbf{W}_L + \Delta \mathbf{W}_{L,t}\}$. 

To keep our notations simple, let us only consider $l$th convolutional layer that has $k\times k$ filters, $C_{in}$ channels for input feature tensor, and $C_{out}$ channels for output feature tensor. We represent the corresponding $\mathbf{W}_l$ as a tensor of size $k^2 \times C_{in} \times C_{out}$. 
We represent the low-rank tensor update as a summation of $R$ rank-1 tensors as 
\begin{equation}
    \label{eq:low_rank_update}
    \Delta \mathbf{W}_{l,t} = \sum^{R}_{r=1} \mathbf{w}^{r}_{1,t} \otimes \mathbf{w}^{r}_{2,t} \otimes \mathbf{w}^{r}_{3,t},
\end{equation}
where $\mathbf{w}^r_{1,t},\mathbf{w}^r_{2,t},\mathbf{w}^r_{3,t}$ represent vectors of length $k^2,C_{in},C_{out}$, respectively, and $\otimes$ represents the Kronecker product. 

Apart from low-rank tensor update, we also optimize over Batch Normalization layers (BN) for each task/domain \cite{ioffe2015batch,pham2021continual}. The BN layer  learns two vectors $\Gamma$ and $\beta$, each of length $C_{out}$. The BN operation along $C_{out}$ dimension can be defined as element-wise multiplication and addition:
\begin{equation}
\text{BN}_{\Gamma, \beta} (u) = \Gamma \left(\frac{u-\mathbb{E}[u]}{ \sqrt{\mathrm{Var}[u] + \epsilon}}\right) + \beta. 
\end{equation} 
We represent the output of $l$th convolutional layer for task/domain $t$ as 
\begin{equation}
    \mathbf{Z}_{l,t} = \text{BN}_{\Gamma_t, \beta_t} (\text{conv}(\mathbf{W}_l + \Delta \mathbf{W}_{l,t}, \mathbf{Y}_{l-1,t})),
\end{equation}
where $\mathbf{Y}_{l-1,t}$ represents the input tensor and $\mathbf{Z}_{l,t}$ represents the output tensor for $l$th layer. 
In our proposed FTN, we learn the task/domain-specific factors $\{\mathbf{w}^r_{1,t},\mathbf{w}^r_{2,t},\mathbf{w}^r_{3,t}\}_{r=1}^R$, and $\Gamma_t$, and $\beta_t$ for every layer in the backbone network.






In the FTN method, rank $R$ for $\Delta \mathbf{W}$  plays an important role in defining the expressivity of the adapted network. We can define a complex $\Delta \mathbf{W}$ by increasing the rank $R$ of the low-rank tensor and taking their linear combination. Our experiments showed that this has resulted in a significant performance gain.

\noindent
\textbf{Initialization.} 
To establish a favorable starting point, we adopt a strategy that minimizes substantial modifications to the frozen backbone network weights during the initialization of the task-specific parameter layers. To achieve this, we initialize each parameter layer from the Xavier uniform distribution \cite{glorot2010understanding}, thereby generating $\Delta \mathbf{W}$ values close to 0 before their addition to the frozen weights. This approach ensures  the initial point of our proposed network closely matches the pretrained network closely. \  

To acquire an effective initialization for our backbone network, we leverage the pretrained weights obtained from ImageNet. We aim to establish a robust and capable feature extractor for our specific task by incorporating these pretrained weights into our backbone network. 

\noindent\textbf{Number of parameters.} In a Fine-Tuning setup with $T$ tasks/domains, the total number of required parameters at convolutional layer $l$ can be calculated as $T \cdot (k^2 \times C_{in} \times C_{out})$. Whereas using our proposed FTNs, the total number of frozen backbone $(\mathbf{W}_l)$ and low-rank R tensor $(\Delta \mathbf{W}_{l,t})$ parameters are given by $(C_{out} \times C_{in} \times k^2) + T \cdot R \cdot (C_{out}+C_{in}+k^{2})$. In our results section, we have shown that the absolute number of parameters required by our method is a fraction of what the Fine-Tuning counterpart needs.

\noindent\textbf{Effect of Batch Normalization.} In our experiment section, under the `FC and BN only' setup, we have shown that having task-specific Batch Normalization layers in the backbone network significantly affects the performance of a downstream task/domain. For all the experiments with our proposed approach, we include Batch Normalization layers as task-specific along with low-rank tensors and classification/decoder layer.

\subsection{FTN applied to Transformers}

The Vision Transformer (ViT) architecture  \cite{dosovitskiy2020image} consists a series of MLP, normalization, and Multi-Head Self-Attention (MHSA) blocks. The MHSA blocks perform $n$ parallel attention mechanisms on sets of Key $K$, Query $Q$, and Value $V$ matrices. Each of these matrices has dimensions of $S \times d_{model}$, where $d_{model}$ represents the embedding dimension of the transformer, and $S$ is the sequence length. The $i$-th output head ($H_i$) of the $n$ parallel attention blocks is computed as
\begin{equation}
H_i = \text{SA}(Q\mathrm{W}_i^Q, K\mathrm{W}_i^K, V\mathrm{W}_i^V),
\end{equation}
where SA($\cdot$) represents the self-attention mechanism, $\mathrm{W}_i^K, \mathrm{W}_i^Q, \mathrm{W}_i^V \in \mathbb{R}^{d_{model} \times d}$ represent the projection weights for the key, query, and value matrices, respectively, and $d = d_{model}/n$. The heads $H_i$ are then combined using a projection matrix $\mathrm{W}_o \in \mathbf{R}^{d_{model}\times d_{model}}$ to result in the output of the MHSA block as
\begin{equation}
\text{MHSA}(H_1,\dots,H_n) = \text{Concat}(H_1, \ldots, H_n) \cdot \mathrm{W}_{o}.
\end{equation}
%
Following the adaptation procedure in \cite{he2023parameter}, we apply our proposed factorization technique to the weights in the MHSA block. We introduce two methods for applying low-rank tensors to the attention weights: \\
\noindent \textbf{Adapting query and value weights. } Our first proposed method, \emph{FTN (Query and Value)}, adds the low-rank tensor factors to the query $\mathrm{W}^Q$ and value $\mathrm{W}^V$ weights. These weights can be represented as three-dimensional tensors of size $d_{model} \times d \times n$. Using \eqref{eq:low_rank_update}, we can define and learn low-rank updates $\Delta \mathbf{W_q}$ and $\Delta \mathbf{W_v}$ for the query and value weights, respectively. \\
\noindent \textbf{Adapting output weights.} Our second method, \emph{FTN (Output projection)}, adds low-rank factors, $\Delta \mathbf{W}_o$, to the output projection weights $\mathbf{W}_o \in \mathbf{R}^{d_{model}\times d \times n}$. Similar to the previous low-rank updates, the updates to the output weights defined following \eqref{eq:low_rank_update}. 

\noindent\textbf{Initialization.} We initialize each low-rank factor by sampling from a Gaussian distribution with $\mu = 0$ and $\sigma = 0.05$. This ensures near-zero initialization, closely matching the pretrained network. 

\noindent\textbf{Number of parameters.} 
The total number of parameters needed for $R$ low-rank tensors and $L$ MHSA blocks in FTN (Query and Value) is $2LR(d_{model}+d+n)$. FTN (Output Projection) requires only $LR(d_{model}+d+n)$ to add a similar number of factors. These additional parameters are significantly fewer than the parameters required for fully fine-tuning the four attention weights, which equals $4Ld_{model}^2$. When compared to other parameter-efficient adaptation methods such as LoRA \cite{hu2021lora} and KAdaptation \cite{he2023parameter}, our methods show superior parameter efficiency. 
The primary distinction is in the method of weight factorization and decomposition. In LoRA, to introduce rank $R$ factors in the query and value weight matrices, $4LRd_{\text{model}}$ parameters are required. Our approach begins with a three-dimensional representation of the attention weights, sized $d_{\text{model}} \times d \times n$. We chose this approach because it allows us to exploit the relationship between the attention heads, further improving parameter efficiency. Moreover, we have explored different types of updates within the self-attention mechanism and proposed two variants of our FTN (\emph{Query and Value} and \emph{Output projection}).
SSF \cite{lian2022scaling} requires $mLd_{model}$, where $m$ is the number of SSF modules in each transformer layer. In Table \ref{tab:ftn_trans_res}, we report the exact number of parameters and demonstrate that our proposed method, \emph{FTN (Output Projection)}, has the best parameter efficiency.

%% file: sections/experiment.tex
\section{Experiments and Results}

\begin{table*}[t]
    \centering
    \small
    \caption{Number of parameters and top-1\% accuracy for baseline methods, comparative methods, and FTN with varying ranks on the five domains of the ImageNet-to-Sketch benchmark experiments.  Additionally, the mean top-1\% of each method across all domains is shown. The `Params' column gives the number of parameters used as a multiplier of those for the Feature-Extractor method, along with the absolute number of parameters required in parentheses. \add{\textbf{Bold} and \underline{underline} indicate the best and second-best results, respectively.}}
    \label{tab:imagenet_sketch}
    \resizebox{0.75\linewidth}{!}{%
    \begin{tabular}{llllllll}
    \toprule
         \textbf{Methods} & \textbf{Params (Abs)} & \textbf{Flowers} & \textbf{Wikiart} & \textbf{Sketch} & \textbf{Cars} & \textbf{CUB} & \textbf{mean} \\
         \midrule
         Fine-Tuning & $6\times$ \;(141M) & 95.69 & \textbf{78.42} & \textbf{81.02} & 91.44 & \underline{83.37} & \textbf{85.98} \\
         Feature-Extractor & $1\times$\; (23.5M) & 89.57 & 57.7 & 57.07 & 54.01 & 67.20 & 65.11\\
         FC and BN only & $1.001\times$\; (23.52M) & 94.39 & 70.62 & 79.15 & 85.20 & 78.68 & 81.60\\
         \midrule
         Piggyback {\cite{mallya2018piggyback}} & $6\times$\;[$2.25\times$] (141M) & 94.76 & 71.33 & 79.91 & 89.62 & 81.59 & 83.44\\
         Packnet $\rightarrow$ {\cite{mallya2018packnet}} & [$1.60\times$] (37.6M) & 93 & 69.4 & 76.20 & 86.10 & 80.40 & 81.02\\
         Packnet $\leftarrow$ {\cite{mallya2018packnet}} & [$1.60\times$] (37.6M)& 90.60 & 70.3 & 78.7 & 80.0 & 71.4 & 78.2\\
         Spot-Tune {\cite{guo2019spottune}} & $7\times$\;[$7\times$] (164.5M) & 96.34 & 75.77 & 80.2 & \textbf{92.4} & \textbf{84.03} & 85.74\\
         WTPB {\cite{mancini2018adding}}  & $6\times$\;[$2.25\times$] (141M) & 96.50 & 74.8 & 80.2 & 91.5 & 82.6 & 85.12\\
         BA$^2$ {\cite{berriel2019budget}} & $3.8\times$ \;[$1.71\times$] (89.3M) & 95.74 & 72.32 & 79.28 & \underline{92.14} & 81.19 & 84.13\\
         TAPS {\cite{wallingford2022task}} & $4.12\times$ (96.82M) & \textbf{96.68} & 76.94 & \underline{80.74} & 89.76 & 82.65 & 85.35\\
         \midrule
         \textbf{FTN, R=1} & $1.004\times$\;(23.95M) & 94.79 & 73.03 & 78.62 & 86.85 & 80.86 & 82.83 \\
         \textbf{FTN, R=50} & $1.53\times$\;(36.02M) & \underline{96.42} & \underline{78.01} & 80.6 & 90.83 & 82.96 & \underline{85.76} \\
         \bottomrule
    \end{tabular} }
\end{table*}




We evaluated the performance of our proposed FTN on several MTL/MDL datasets. We performed experiments for \textbf{1. Multi-domain classification} on convolution and transformer-based networks, and \textbf{2. Multi-task dense prediction}.  For each set of benchmarks, we reported the performance of FTN with different rank increments and compared the results with those from existing methods. All experiments are run on a single NVIDIA GeForce RTX 2080 Ti GPU with 12GB memory.

\subsection{Multi-domain classification}
\subsubsection{Convolution-based networks} 


\noindent\textbf{Datasets.} We use two MTL/MDL classification-based benchmark datasets. First, ImageNet-to-Sketch, which contains five different domains: Flowers, Cars, Sketch, Caltech-UCSD Birds (CUBs), and WikiArt, with different classes. Second, DomainNet, which contains six domains: Clipart, Sketch, Painting (Paint), Quickdraw (Quick), Inforgraph (Info), and Real, with each domain containing an equal 345 classes. The datasets are prepared using augmentation techniques as adopted by~\cite{wallingford2022task}.

\noindent \textbf{Training details.} For each benchmark, we report the performance of FTN for various choices for ranks, along with several benchmark-specific comparative and baseline methods. The backbone weights are pretrained from ImageNet, using ResNet-50 for the ImageNet-to-Sketch benchmarks, and ResNet-34 on the DomainNet benchmarks to keep the same setting as~\cite{wallingford2022task}. On ImageNet-to-Sketch we run FTNs for ranks, $R\in\{1,5,10,15,20,25,50\}$ and on DomainNet dataset for ranks, $R\in\{1,5,10,20,30,40\}$. In the supplementary material, we provide the hyperparameter details to train FTN. 


\noindent \textbf{Results.} We report the top-1\% accuracy for each domain and the mean accuracy across all domains for each collection of benchmark experiments.  We also report the number of frozen and learnable parameters in the backbone network. Table~\ref{tab:imagenet_sketch} compares the FTN method with other methods in terms of accuracy and number of parameters.
FTN outperforms every other \add{adaptation-based} method \add{in number of parameters} while using 36.02 million parameters in the backbone with rank-50 updates for all domains. The mean accuracy performance is better than other \add{adaptation-based} methods and is close to Spot-Tune \cite{guo2019spottune} \add{ and Fine-Tuning}, which requires nearly 165M \add{and 141M }parameters \add{respectively}. On the Wikiart dataset, we outperform the top-1 accuracy with other \delete{baseline} \add{adaptation-based} methods. The performance of baseline methods is taken from TAPS \cite{wallingford2022task}, as we are running the experiments under the same settings.

\begin{table*}[tbhp]
\centering
    \small
    \caption{Performance of different methods with resnet-34 backbone on DomainNet dataset. Top-1\% accuracy is shown on different domains with different methods along with the number of parameters. \add{\textbf{Bold} and \underline{underline} indicate the best and second-best results, respectively.}}
    \label{tab:domainnet_incremental}
    \resizebox{0.75\linewidth}{!}{%
        \begin{tabular}{lllllllll}
        \toprule
             \textbf{Methods} & \textbf{Params} & \textbf{Clipart} & \textbf{Sketch} & \textbf{Paint} & \textbf{Quick} & \textbf{Info} & \textbf{Real} & \textbf{mean} \\
             \midrule
             Fine-Tuning & $6\times$  & 74.26 & \textbf{67.33} & 67.11 & \textbf{72.43} & \textbf{40.11} & \underline{80.36} & \textbf{66.93} \\
             Feature-Extractor & $1\times$  & 60.94 & 50.03 & 60.22 & 54.01 & 26.19 & 76.79 & 54.69 \\
             FC and BN only & $1.004\times$ & 70.24 & 61.10 & 64.22 & 63.09 & 34.76 & 78.61 & 62.00 \\
             \midrule
             Adashare {\cite{sun2020adashare}} & $5.73\times$ & \underline{74.45} & 64.15 & 65.74 & 68.15 & 34.11 & 79.39 & 64.33 \\
             TAPS {\cite{wallingford2022task}} & $4.90\times$ & \textbf{74.85} & \underline{66.66} & \textbf{67.28} & \underline{71.79} & 38.21 & 80.28 & \underline{66.51} \\
             \midrule
             \textbf{FTN, R=1} & $1.008\times$ & 70.73 & 62.69 & 65.08 & 64.81 & 35.78 & 79.12 & 63.03 \\
             \textbf{FTN, R=40} & $1.18\times$ & 74.2 & 65.67 & \underline{67.14} & 71.00 & \underline{39.10} & \textbf{80.64} & 66.29 \\
             \bottomrule
        \end{tabular}}
\end{table*}

Table~\ref{tab:domainnet_incremental} shows the results on the DomainNet dataset, which we compare with TAPS \cite{wallingford2022task} and Adashare \cite{sun2020adashare}. Again, using FTN, we significantly outperform comparison methods along the required parameters (rank-40 needs ~25.22 million parameters only). Also, FTN rank-40 attains better top-1\% accuracy on the Infograph and Real domain, while it attains similar performance on all other domains. On DomainNet with resnet-34 and Imagenet-to-Sketch with resnet-50 backbone, the rank-1 low-rank tensors require only 16,291 and 49,204 parameters per task, respectively. We have shown additional experiments on this dataset under a joint optimization setup in section 4 of the supplementary material.

\begin{figure*}[!t]
     \centering
     \begin{subfigure}[t]{0.4\textwidth}
         \centering
         \includegraphics[width=\textwidth]{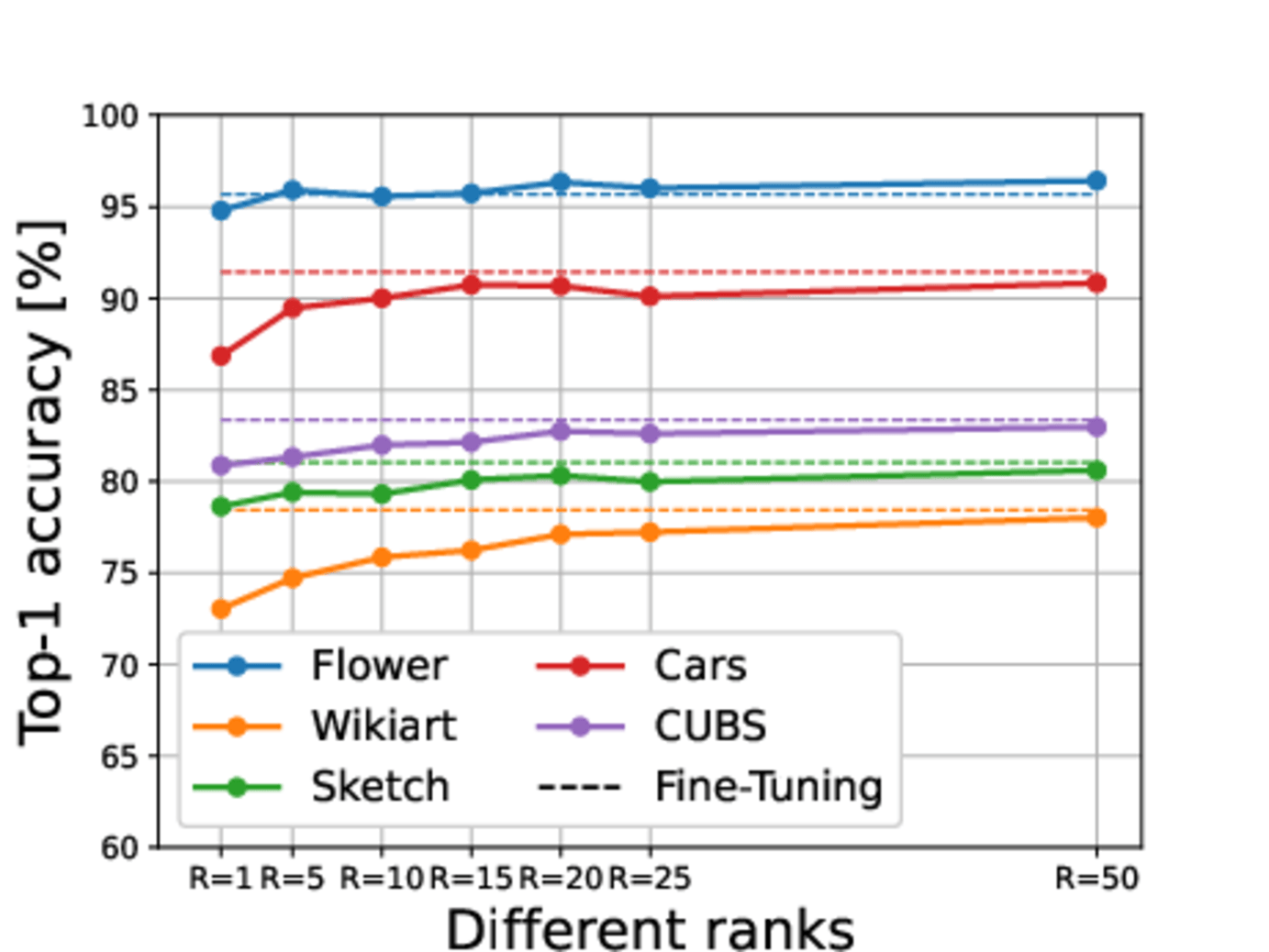}
         \caption{\add{Imagenet-to-sketch dataset}}
         \label{imgnet-to-sketch_rank}
     \end{subfigure}
     \begin{subfigure}[t]{0.4\textwidth}
         \centering
         \includegraphics[width=\textwidth]{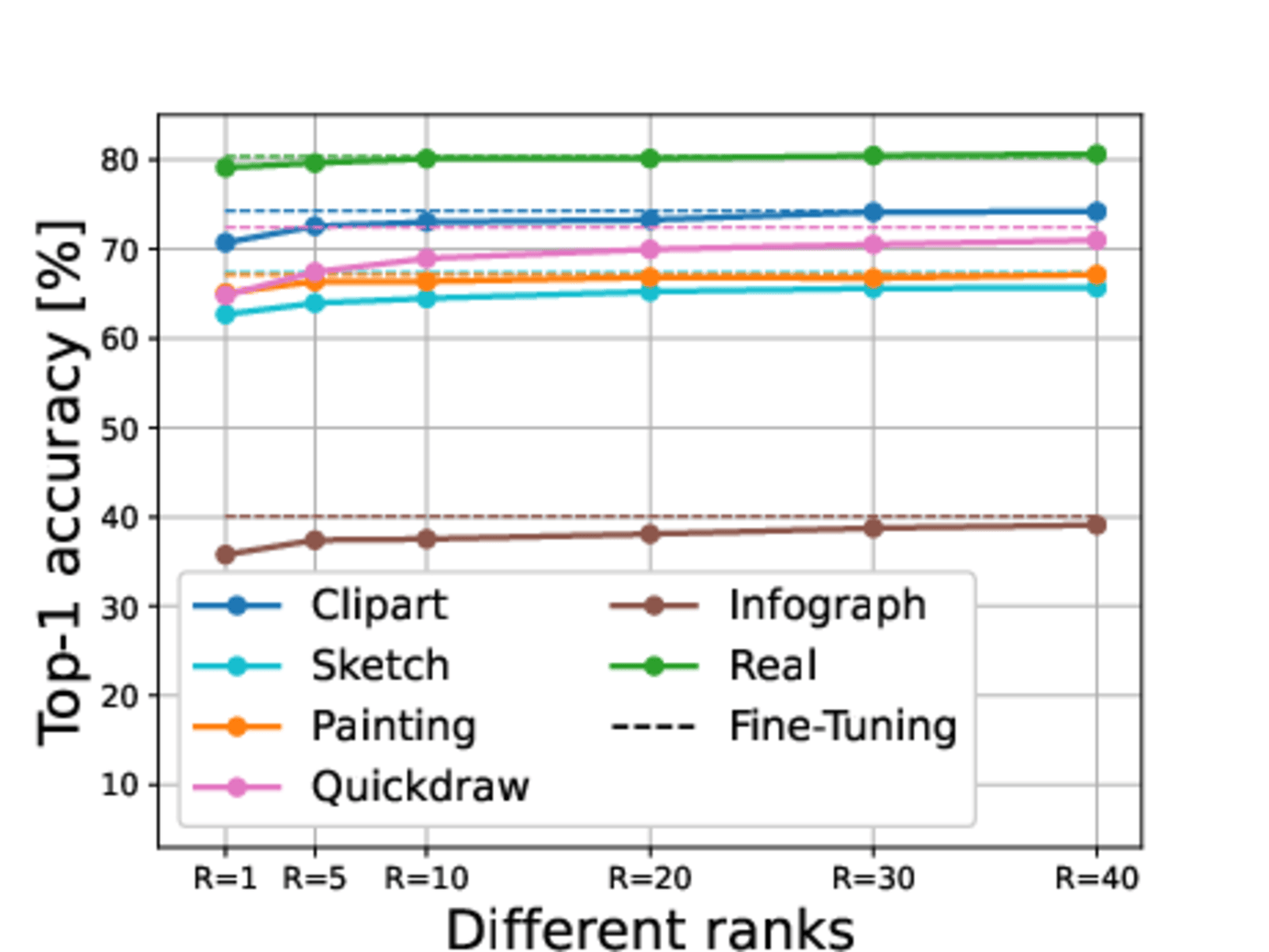}
         \caption{\add{DomainNet dataset}}
         \label{domainnet_rank}
     \end{subfigure}
    \caption{\textbf{Accuracy vs Low-ranks:} We show the top-1\% accuracy against different low-ranks used in our method for different domains. We start with ‘only BN’ setup where without any low-rank we keep the Batch Normalization layers as task-specific. Then we show the performance improvement through our approach upon increasing the rank-R.}
        \label{fig:AccuracyvsLowranks}
\end{figure*}

\noindent\textbf{Analysis on rank.} We create low-rank tensors ($\Delta W$) as a summation of $R$ rank-1 tensors. We hypothesize that increasing $R$ increases the expressive power of low-rank tensors. Our experiments confirm this hypothesis, where increasing the rank improves the performance, enabling more challenging task/domain adaptation. Figure~\ref{fig:AccuracyvsLowranks} shows the accuracy vs. ranks plot, where we observe a trend of performance improvement as we increase the rank from 1 to 50 on the ImageNet-to-Sketch and from 1 to 40 on the DomainNet dataset. In addition, we observe that some domains do not require high ranks. Particularly, the Flowers and Cars domains attain good accuracy at ranks 20 and 15, respectively. We can argue that, unlike prior works \cite{guo2019spottune, li2016revisiting}, which consume the same task-specific module for easy and complex tasks, we can provide different flexibility to each task. Also, we can add as many different tasks as we want by adding independent low-rank factors for each task (with a sufficiently large rank). 
In supplementary material, we present a heatmap that shows the adaption of the low-rank tensor at every layer upon increasing the rank. Section 2 of the supplementary materials shows an additional experiment to demonstrate the effect on performance with different numbers of low-rank factors.

\begin{table*}[!t]
\centering
\small
\caption{We compared performance across five datasets in terms of accuracy and total parameters. FTN (O) uses low-rank factors for output projection weights, while FTN (Q\&V) applies them to query and value weights. Note that the parameters mentioned exclude task-specific heads, and $5 \times (439.5M)$ denotes a fivefold increase from the base network's 87.9M parameters. \add{\textbf{Bold} and \underline{underline} indicate the best and second-best results, respectively.}}
\label{tab:ftn_trans_res}
\resizebox{\linewidth}{!}{%
\begin{tabular}{lcccccccccc}
\toprule
Method                 & Params (Abs) & \# additional params & \add{FLOPS} & \add{Wall-clock time} & CIFAR10 & CIFAR100 & DTD  & STL10 & FER2013 & mean \\ \midrule
Fine-tuning            & $5\times$ (439.5M) &  $5\times$ 87.9M & \add{\textbf{4.368G}} & \add{168.62} & \textbf{97.7}    & \textbf{85.4}     & \textbf{79.0} & \textbf{99.7}  & \textbf{69.8}    & \textbf{86.3}         \\
Feature extractor      & $1\times$ (87.9M) & - & \add{\textbf{4.368G}} & \add{\textbf{81.40}} & 94.8    & 80.1     & 75.4 & 98.4  & 67.3    & 83.2         \\
LoRA  {\cite{hu2021lora}}  & $1.008\times$ (88.6M) & $5\times $147.2K & \add{\underline{4.421G}} & \add{164.83}  & 95.1    & \underline{78.1}     & 78.1 & \underline{99.2}  & 67.7    & 83.6         \\
KAdaptation {\cite{he2023parameter}}           & $1.005\times$ (88.3M) & $5\times $80.7K & \add{5.349G} & \add{158.69}  & 95.9    & \underline{84.8}     & \underline{78.1} & \underline{99.2}  & 69.0    & \underline{85.4}         \\
$ \text{SVFT}^{P} $ {\cite{lingam2024svft}}    & $1.003\times$ (88.1M) & $5\times $55.3K & \add{69.59G} & \add{221.21}  & \underline{97.1}    & 83.6     & 73.3 & 98.1  & \underline{69.5}    & 84.3         \\ \midrule
\textbf{FTN (Q \& V)}  & $1.005\times$ (88.3M) & $5\times$  81.0K & \add{5.178G} & \add{146.89}& 95.8    & 83.4     & 77.1 & 98.7  & 68.5    & 84.7         \\
\textbf{FTN (O)} & ${1.002\times}$ (88.1M) & $5\times$40.5K & \add{4.442G} & \add{\underline{129.82}} & 96.6    & 84.3     & 76.0 & 98.6  & \underline{69.5}    & 85.0         \\ \bottomrule
\end{tabular}%
}
\end{table*}
\subsubsection{Transformer-based networks}\

\noindent We compared our FTN method with several domain adaptation techniques for supervised image classification. Our task is to adapt a pretrained 12-layer ViT-B-224/32 (CLIP) model obtained from \cite{he2023parameter} to new domains. 

\noindent\textbf{Datasets.} We conducted experiments on the CIFAR10, CIFAR100, DTD, FER2013, and STL10 classification datasets, using the official dataset splits. 

\noindent\textbf{Training details.} For all experiments except SVFT \cite{lingam2024svft}, we set the rank to $R=4$. We followed a similar hyper-parameter tuning procedure and implementation as outlined in \cite{he2023parameter}, which utilizes grid-search to obtain the optimal learning rate for each dataset. We found that $5\times 10^{-6}$ was the optimal learning rate. Following the approach in \cite{hu2021lora}, we scaled the low-rank factors by $\frac{\alpha}{R}$, where $\alpha$ is a hyper-parameter, and $R$ is the number of low-rank factors. We set $\alpha=10$ and $\alpha=100$ for FTN (Query and Value) and FTN (Output projection), respectively. We used a batch size of $64$ and trained for $100$ epochs. For SVFT, we used its Plain variant from their codebase to maintain a comparable number of additional parameters and performed hyper-parameter tuning to determine optimal learning rates for a fair comparison.   

\noindent\textbf{Results.} In Table \ref{tab:ftn_trans_res}, we present the classification accuracy and the total number of parameters for our proposed FTN methods, along with related model adaptation methods. Results for Fine-tuning, Feature extractor (Linear-probing), LoRA \cite{hu2021lora}, and KAdaptation \cite{he2023parameter} are obtained from \cite{he2023parameter}. The first proposed method, FTN (query and value), surpasses LoRA in terms of average performance and requires fewer additional parameters. FTN (query and value) requires a comparable number of parameters to KAdaptation and performance is $0.8\%$ lower. In contrast, FTN (output projection) requires approximately half as many additional parameters as KAdaptation but achieves comparable performance. Additionally, FTN outperforms SVFT \cite{lingam2024svft} on average while using fewer parameters. \add{Fine-tuning and Feature extractor methods require the least FLOPS due to the absence of architectural modifications. Among the others, LoRA and FTN (O) achieve comparable and second-best FLOPS. We calculate Wall-clock time as the total duration, in seconds, required to complete a single training epoch. The Feature extractor
approach had the shortest wall-clock time, as expected due to the frozen backbone. FTN (O) achieves the best wall-clock performance among the remaining methods, highlighting its training efficiency.}
\subsection{Multi-task dense prediction}
\noindent
\textbf{Dataset.} The widely-used NYUD dataset with 795 training and 654 testing images of indoor scenes is used for dense prediction experiments in multi-task learning. The dataset contains four tasks: edge detection (Edge), semantic segmentation (SemSeg), surface normals estimation (Normals), and depth estimation (Depth). We follow the same data-augmentation technique as used by \cite{kanakis2020reparameterizing}. 

\noindent\textbf{Metrics.} On the tasks of the NYUD dataset, we report mean intersection over union for semantic segmentation, mean error for surface normal estimation, optimal dataset F-measure~\cite{martin2004learning} for edge detection, and root mean squared error for depth estimation.  We also report the number of parameters used in the backbone for each method. 


\noindent\textbf{Training details.} ResNet-18 is used as the backbone network, and DeepLabv3+ as the decoder architecture. The Fine-Tuning and Feature-Extractor experiments are implemented in the same way as in the classification-based experiments above. We showed experiments for FTNs with $R\in\{1,10,20,30\}$. Further details are in the supplementary material. 

\noindent \textbf{Results.} Table~\ref{tab:nyud} shows the performance of FTN with various ranks and of other baseline comparison methods for dense prediction tasks on the NYUD dataset.  We observe performance improvement by increasing flexibility through higher rank. FTN with rank-30 performs better than all comparison methods and utilizes the least number of parameters. Also, we attain good performance on the Depth and Edge task by using only rank-20. We take the performance of baseline comparison methods from the RCM paper \cite{kanakis2020reparameterizing} as we run our experiments under the same setting. 

\noindent Section 6 of the supplementary materials presents additional experiments on the multi-domain image generation application using the FTN method.

\begin{table}[h]
    \centering
    \caption{Dense prediction performance on NYUD dataset using ResNet-18 backbone with DeepLabv3+ decoder. The proposed FTN approach with $R=\{1,10,20,30\}$ and other methods. \delete{The best performing method in bold.}\add{\textbf{Bold} and \underline{underline} indicate the best and second-best results, respectively.}}
    \label{tab:nyud}
    \resizebox{\columnwidth}{!}{
    \begin{tabular}{llllll}
        \toprule
        \textbf{Methods}    & \textbf{Params} & \textbf{Semseg$\uparrow$} & \textbf{Depth$\downarrow$} & \textbf{Normals$\downarrow$} & \textbf{Edge$\uparrow$} \\
        \toprule
        Single Task         & $4\times$         & \underline{35.34}           &\underline{0.56}           & 22.20            & 73.5          \\
        Decoder only        & $1\times$           & 24.84           & 0.71           & 28.56            & 71.3          \\
        Decoder + BN only   & $1.002\times$       & 29.26           & 0.61           & 24.82            &  71.3             \\
        \midrule
        ASTMT (R-18) {\cite{maninis2019attentive}} & $1.25\times$   & 30.69           & 0.60           & 23.94            & 68.60         \\
        ASTMT (R-26+SE) {\cite{maninis2019attentive}}    &  $2.00\times$   & 30.07           & 0.63           & 24.32            & 73.50         \\
        Series RA {\cite{rebuffi2018efficient}}          & $1.56\times$  & 31.87           & 0.60           & 23.35            & 67.56         \\
        Parallel RA {\cite{rebuffi2018efficient}}        &  $1.50\times$    & 32.13           & 0.59           & 23.20            & 68.02         \\
        RCM   {\cite{kanakis2020reparameterizing}}             &  $1.56\times$    & 34.20           & 0.57           & 22.41            & 68.44         \\
        \midrule
        \textbf{FTN, R=1}  & $1.005\times$       & 29.83  & 0.60  & 23.56   & 72.7          \\
        \textbf{FTN, R=10} & $1.03\times$      & 33.66  & 0.57  & 22.15   & 73.5          \\
        \textbf{FTN, R=20} & $1.06\times$       & 34.06  &\textbf{0.55}  & \underline{21.84}   &\textbf{73.9}          \\
        \textbf{FTN, R=30} & $1.09\times$       &\textbf{35.46}  & \underline{0.56}  &\textbf{21.78}   & \underline{73.8}    \\     
        \bottomrule
    \end{tabular}}
\end{table}
%

%


%% file: sections/conclusion.tex
\section{Conclusion}
We have proposed a simple, parameter-efficient, architecture-agnostic, and easy-to-implement FTN method that adapts to new unseen domains/tasks using low-rank task-specific tensors. Our work shows that FTN requires the least number of parameters compared to other baseline methods in MDL/MTL experiments and attains better or comparable performance. We can adapt the backbone network in a flexible manner by adjusting the rank according to the complexity of the domain/task. We conducted experiments with different convolutional backbones and transformer architectures for various datasets to demonstrate that FTN outperforms existing methods. 

\noindent
\textbf{Future work.} In our current approach, we used a fixed rank for each layer. Moving forward, we can explore adaptively selecting the rank for different layers, which may further reduce the number of parameters. MDL/MTL models are often challenged by task interference or negative transfer learning when conflicting tasks are trained together. Future work can address this by investigating which tasks or domains should be learned jointly to mitigate such drawbacks. Additionally, while our method requires a separate forward pass for each task due to the shared backbone, we could further explore branched or tree-structured models that enable task-specific layer sharing to reduce latency.

%% file: supplementary.tex
\section*{Supplementary Material}

\setcounter{section}{0}
\renewcommand{\thesection}{S\arabic{section}}
\renewcommand{\thefigure}{S\arabic{figure}}
\renewcommand{\thetable}{S\arabic{table}}

\section{Dataset and training details}
\subsection{Imagenet-to-sketch dataset} The dataset contains five different domains: Flowers \cite{nilsback2008automated}, Cars \cite{krause20133d}, Sketch \cite{eitz2012humans}, Caltech-UCSD Birds (CUBs) \cite{wah2011caltech}, and WikiArt \cite{saleh2016large} with 102, 195, 250, 196, and 200 classes, respectively. We randomly crop images from each domain to $224\times 224$ pixels, along with normalization and random horizontal flipping. We report the baseline experiments, Fine-Tuning, Feature-Extractor, and 'FC and BN only' with 0.005 learning rate (lr) and SGD optimizer with no weight decay. We train for 30 epochs with a batch size 32 and a cosine annealing learning rate scheduler. \ 

The experiments with our proposed FTN approach are learned through the Adam optimizer with lr 0.005 for low-rank layers and through the SGD optimizer with lr 0.008 for remaining trainable layers (task-specific batchnorm and classification layers). Again, we train them for 30 epochs with batch size 32 and cosine annealing scheduler. We showed the experiments for different low-ranks, $R\in\{1,5,10,15,20,25,50\}$.

We additionally compare the proposed FTN under differnet backbone architecture, EfficientNet-B4. I Table~\ref{tab:efficientnet_b4} we have shown the performance for Fine-Tuning, Feature-Extractor, and FTN with for $R\in\{1,10\}$. FTN irrespective of backbone architecture is giving performance close to Fine-Tuning.

\begin{table*}[h]
    \centering
    \small
\caption{Performance on ImageNet-to-Sketch dataset under incremental setting using EfficientNet-B4 backbone with our FTN approach.}
    \label{tab:efficientnet_b4}
    \begin{tabular}{llllllll}
        \toprule
        \textbf{Method}            & \textbf{Params (Abs)}   & \textbf{Flowers} & \textbf{Wikiart} & \textbf{Sketch} & \textbf{Cars}  & \textbf{CUB}   & \textbf{mean}  \\
        \midrule
        Fine-Tuning       & 6x(105.24M)    & 96.08   & 78.72   & 80.9   & 92.81 & 83.67 & 86.43 \\
        Feature extractor & 1x(17.54M)     & 80.91   & 42.37   & 56.3   & 42.97 & 64.77 & 57.46 \\
        \midrule
        \textbf{FTN, R=1}          & 1.079x(18.93M) & 93.28   & 74.07   & 79.77  & 87.93 & 82.82 & 83.57 \\
        \textbf{FTN, R=10}         & 1.474x(25.88M) & 94.79   & 77.54   & 80.7   & 89.71 & 84.70 & 85.48 \\
        \bottomrule
    \end{tabular}
\end{table*}

\subsection{DomainNet dataset} This dataset contains six domains: Clipart, Sketch, Painting (Paint), Quickdraw (Quick), Inforgraph (Info), and Real, with an equal number of 345 classes/categories. We train the baseline experiments, Fine-Tuning, Feature-Extractor, and 'FC and BN only' with 0.005 lr with 0.0001 weight decay and SGD optimizer. Similar to the Imagenet-to-sketch dataset, we apply the same data augmentation techniques and train for 30 epochs with 32 batch size and a cosine annealing learning rate scheduler. \ 

For our experiments with the FTN method, we train the low-rank tensor layers with Adam optimizer and 0.005 lr. The remaining layers were optimized using the SDG optimizer with the same 0.005 learning rate and no weight decay. We train the FTN networks for 30 epochs with the same learning rate scheduler. We showed our experiments for different low-ranks, $R\in\{1,5,10,20,30,40\}$.
\subsection{NYUD dataset} In multi-task learning, we use NYUD dataset, which consists of 795 training and 654 testing images of indoor scenes, for dense prediction experiments. It has four tasks: edge detection (Edge), semantic segmentation (SemSeg), surface normals estimation (Normals), and depth estimation (Depth). We evaluated the performance using optimal dataset F-measure (odsF) for edge detection, mean intersection over union (mIoU) for semantic segmentation, and mean error (mErr) for surface normals. At the same time, we report root mean squared error (RMSE) for depth. We perform random scaling in the range of [0.5, 2.0] and random horizontal flipping for data augmentation and resize each image to $425\times 560$. We train our baseline experiments, Fine-tuning, Feature Extractor, and 'FC and BN only' for 60 epochs with batch size 8 and polynomial learning rate scheduler. We learn the network using SGD optimizer and 0.005 learning rate with 0.9 momentum and 0.0001 weight decay. \ 

In FTN we train for the same 60 epochs, batch size 8, and polynomial learning rate scheduler. We learn over low-rank layers using the Adam optimizer with a 0.01 learning rate and no weight decay. The remaining decoder and batchnorm layers are optimized using the same hyperparameters used for baseline experiments. This dataset shows experiments with different low-ranks, $R\in\{1,10,20,30\}$.

\subsection{Visual Decathlon Challenge dataset}
We showed the performance of FTN on the visual decathlon challenge dataset. The experiments are performed on resnet-18 backbone with pre-trained weights initialized from Imagenet. In Table~\ref{tab:visual_decathlon}, we report the performance using FTN with rank R=1, 20 and comparison with Fine-Tuning and Feature-Extractor. We also report the total number of parameters used by each method. The FTN with rank, R=20 achieves accuracy significantly better than Feature-Extractor (and comparable to Fine-Tuning networks) while using only $\sim ~6\%$ additional parameters compared to Feature-Extractor.

\begin{table*}[!ht]
    \centering
        \small
    \caption{Performance on Visual Decathlon Challenge dataset using resnet-18 backbone with FTN approach using R=1,20 along with the number of parameters.}
    \label{tab:visual_decathlon}
    \resizebox{0.8\linewidth}{!}{%
    \begin{tabular}{llllllllllll}
    \toprule
    \textbf{Method}   & \textbf{Params (Abs)} & \textbf{Airc.} & \textbf{C100} & \textbf{DPed} & \textbf{DTD} & \textbf{GTSR} & \textbf{Flwr} & \textbf{OGlt} & \textbf{SVHN} & \textbf{UCF} & \textbf{mean} \\
    \midrule
    Fine-tuning       & 10x(111.7M)           & 49.74          & 80.08         & 99.91         & 53.19        & 99.93         & 84.60         & 88.37         & 95.43         & 80.63        & 81.32         \\
    Feature extractor & 1x(11.17M)            & 15.69          & 59.52         & 87.55         & 45.03        & 81.34         & 59.31         & 44.73         & 39.50         & 33.14        & 51.75         \\
    \midrule
    FTN, R=1          & 1.005x(11.23M)        & 31.29          & 77.38         & 99.36         & 48.98        & 99.83         & 65.98         & 85.12         & 93.06         & 55.94        & 72.99         \\
    FTN, R=20         & 1.06x(11.89M)         & 42.93          & 77.41         & 99.82         & 52.97        & 99.94         & 76.96         & 86.12         & 94.6          & 65.52        & 77.36 \\
    \bottomrule
    \end{tabular}}
\end{table*}

\section{Effect on performance with different number of low-rank factors.}
\begin{figure*}[t]
    \small
     \centering
     \begin{subfigure}{0.32\textwidth}
         \centering
         \includegraphics[width=\linewidth]{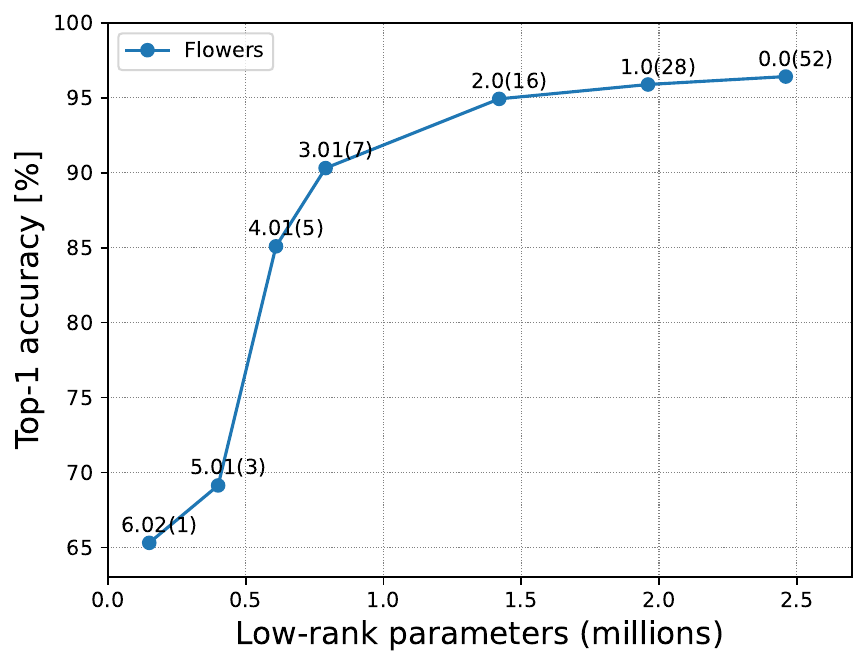}
     \end{subfigure}
     \hfill
     \begin{subfigure}{0.32\textwidth}
         \centering
         \includegraphics[width=\linewidth]{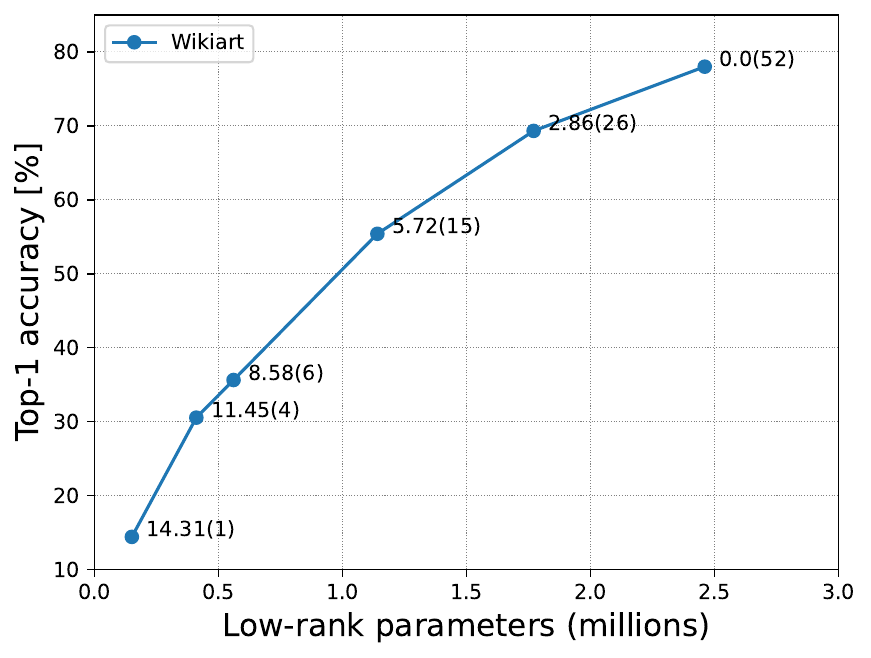}
     \end{subfigure}
     \hfill
     \begin{subfigure}{0.32\textwidth}
         \centering
         \includegraphics[width=\linewidth]{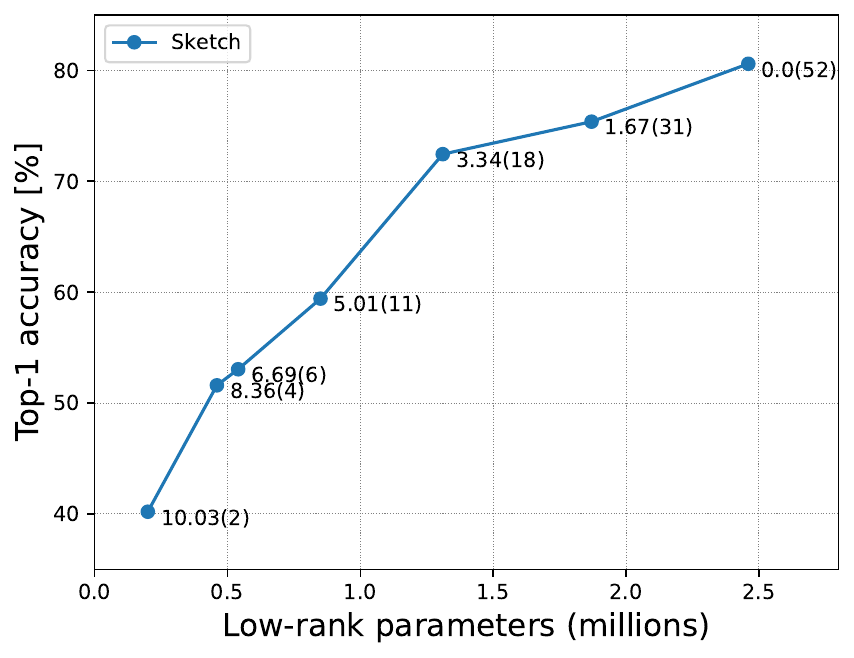}
     \end{subfigure}

     \vspace{0.5cm}
     
     \begin{subfigure}{0.32\textwidth}
         \centering
         \includegraphics[width=\linewidth]{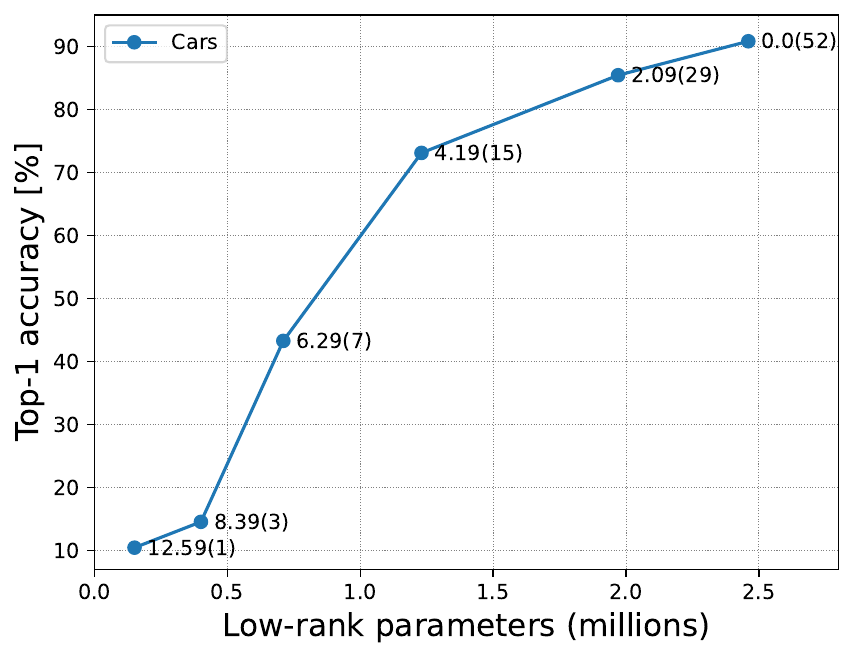}
     \end{subfigure}
     \begin{subfigure}{0.32\textwidth}
         \centering
         \includegraphics[width=\linewidth]{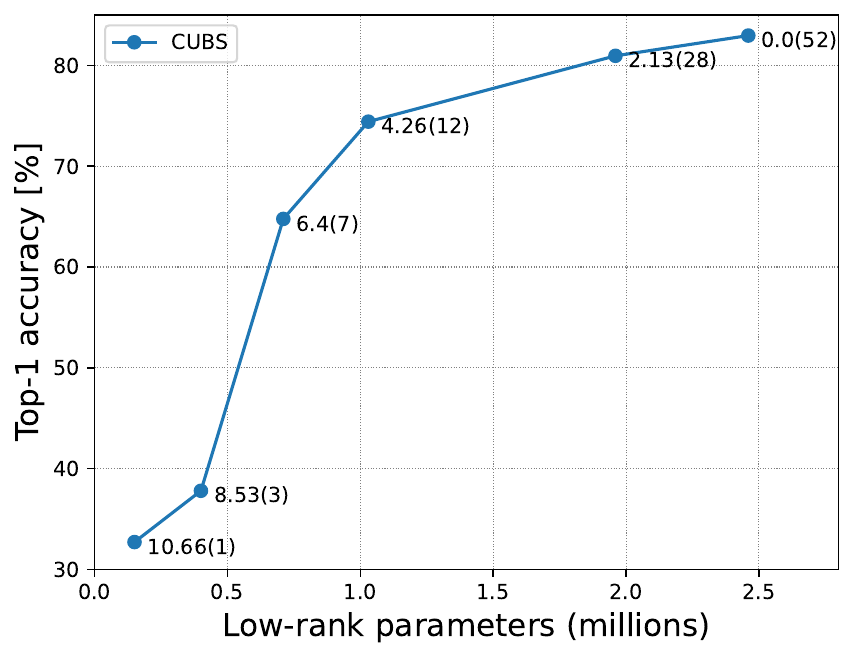}
     \end{subfigure}
     
        \caption{Performance on five domains of the Imagenet-to-sketch dataset as we remove the low-rank parameters. We selected the number of layers in the backbone based on a moving threshold. We annotate the specified threshold at each marker point and the number of affected layers (in parentheses).}
        \label{fig:img_to_sketch_threshold}
         
        \vspace{-2ex}
\end{figure*}

We performed an experiment by removing the low-rank factors from our trained FTN backbone network at different thresholds. We perform this experiment on five domains of the Imagenet-to-sketch dataset and compute the $\ell_2$-norm of $\Delta \mathbf{W}$ at every layer. We selected equally spaced threshold values from the minimum and maximum $\ell_2$-norm and removed the low-rank factors below the threshold. The performance vs. the number of parameters of low-rank layers for different thresholds is shown in Figure~\ref{fig:img_to_sketch_threshold}. We observe a drop in performance on every domain as we increase the threshold and reduce the number of layers from the backbone. Interestingly, when we reduce the number of layers from 52 to 28 on the Flowers and CUBS dataset, we did not see a significant drop in accuracy.

\section{Visualization of changes in FTN with low-rank factors}
\begin{figure*}[!h]
  \centering
  \includegraphics[width=\textwidth]{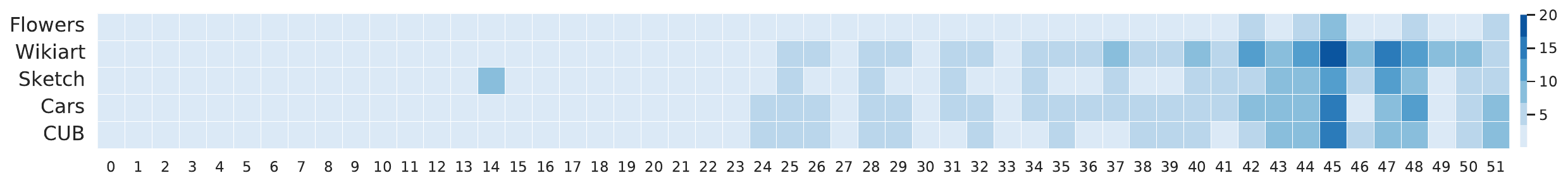}
  \caption{Norm of low-rank factors in the adapted backbone layers for different domains of the Imagenet-to-sketch dataset with $R=50$.}
  \label{fig:heatmap_img_sketch_rank50}
\end{figure*}

\begin{figure*}[!h]
  \centering
  \includegraphics[width=\textwidth]{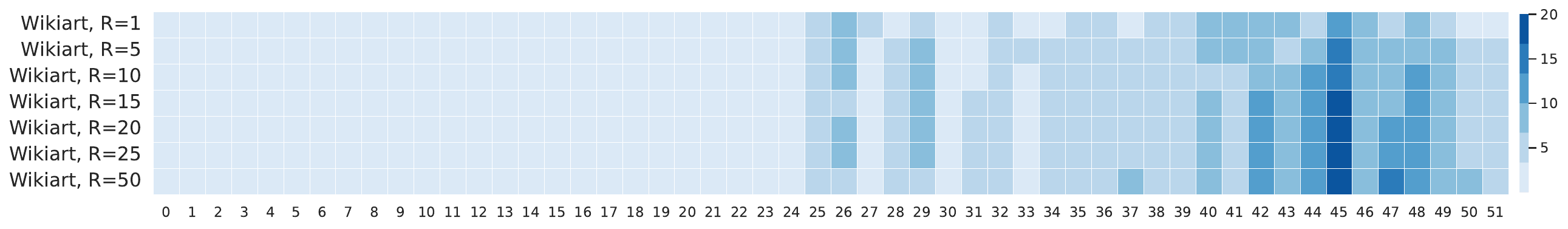}
  \caption{Norm of low-rank factors in the adapted backbone layers for different values of $R\in \{1,5,10,15,20,25,50\}$ with the wikiart domain of the Imagenet-to-sketch dataset. }
  \label{fig:heatmap_wikiart_rank1_50}
\end{figure*}

\begin{figure*}[!h]
  \centering
  \includegraphics[width=\textwidth]{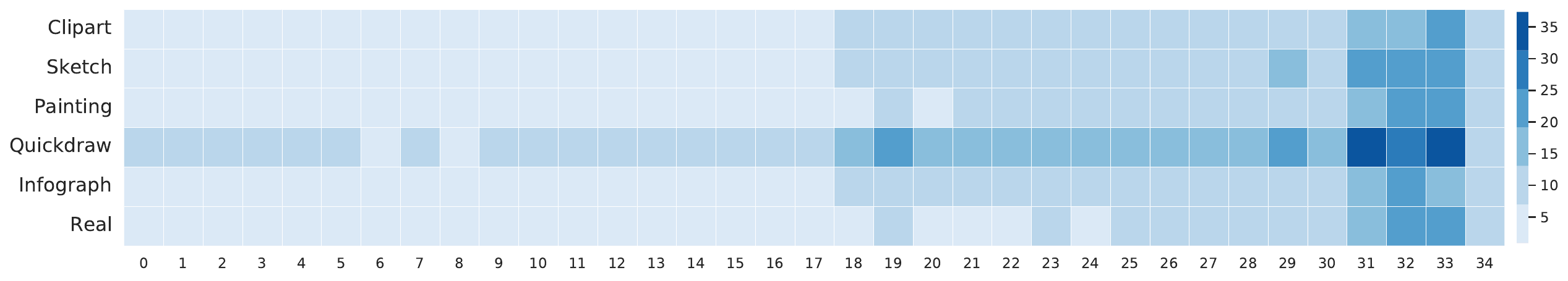}
  \caption{Norm of low-rank factors in the adapted backbone layers for different domains of the DomainNet dataset with $R=40$.}
  \label{fig:heatmap_domainnet_rank40}
\end{figure*}

We present the norm of low-rank factors at every adapted layer in the backbone of our FTN as a heatmap in Figures~\ref{fig:heatmap_img_sketch_rank50}--\ref{fig:heatmap_wikiart_rank1_50}. The colors indicate relative norms because we normalized them for every network in the range 0 to 1 to highlight the relative differences. Figure~\ref{fig:heatmap_img_sketch_rank50} presents results on five domains of the Imagenet-to-sketch dataset (resnet-50 backbone), adapting every layer with rank-50 FTN. We observe the maximum changes in the last layer of the backbone network instead of the initial layers. We also show a similar trend on the DomainNet dataset with resnet-34 backbone where maximum changes occur in the network's later layer (see Figure~\ref{fig:heatmap_domainnet_rank40}). We observe from Figure~\ref{fig:heatmap_wikiart_rank1_50} that on the wikiart domain of the Imagenet-to-sketch dataset, the layers in the backbone network become more adaptive upon increasing the rank of FTN. The rank-50 FTN has more task adaptive layers than the rank-1 FTN on the wikiart domain.

\begin{table*}[htbp]
    \centering
    \small
\caption{Performance on DomainNet dataset under joint setting using resnet-34 backbone (initialized with jointly trained weights) with our FTN approach along with comparison methods.}
    \label{tab:domainnet_jointly}
    \resizebox{0.75\textwidth}{!}{
    \begin{tabular}{lllllllll}
    \toprule
         \textbf{Methods} & \textbf{Params (Abs)} & \textbf{Clipart} & \textbf{Sketch} & \textbf{Paint} & \textbf{Quick} & \textbf{Info} & \textbf{Real} & \textbf{mean} \\
         \midrule
         Fine-tuning & $6\times$ (127.68M) & 77.43 & 69.25 & 69.21 & 71.61 & 41.50 & 80.74 & 68.29 \\
         \add{Feature extractor(unified head)} &  \add{$1\times$ (21.28M)} & \add{33.74} & \add{73.97} & \add{73.13} & \add{54.75} & \add{39.49} & \add{63.64} & \add{56.45} \\
         Feature extractor & $1\times$ (21.28M) & 76.67 & 65.2 & 65.26 & 52.97 & 35.05 & 76.08 & 61.87 \\
         FC and BN only & $1.004\times$ (21.35M) & 77.07 & 68.34 & 68.76 & 69.06 & 40.63 & 79.07 & 67.15 \\
         \midrule
         Adashare \add{\cite{sun2020adashare}} & $1\times$ (21.28M) & 75.88 & 63.96 & 67.90 & 61.17 & 31.52 & 76.90 & 62.88 \\
         TAPS \add{\cite{wallingford2022task}} & $1.46\times$ (31.06M) & 76.98 & 67.81 & 67.91 & 70.18 & 39.30 & 78.91 & 66.84 \\
         \midrule
         \textbf{FTN, R=1} & $1.008\times$ (21.45M) & 77.13 & 68.10 & 68.50 & 69.41 & 40.04 & 79.49 & 67.11 \\
         \bottomrule
         
    \end{tabular} 
    }
    
\end{table*}
 
\section{Results on DomainNet dataset under joint setting}
We performed additional experiments under a joint learning setup on the DomainNet dataset. A single model is trained jointly on all the domains of the dataset with Fine-Tuning setup. Table~\ref{tab:domainnet_jointly} summarizes the results for the performance of the DomainNet dataset under a joint setup. The domains under this setting share information among each other for all parameters. The Fine-Tuning experiment achieves the overall best performance, but at the expense of a large number of parameters. We observe poor performance for the shared Feature Extractor, since that does not learn any additional task-specific parameters. 
\add{For Feature Extractor with unified head, we combined all six domains into a single dataset and then trained a network that uses a single backbone and classifier. From the second row, we observe significant performance improvements in domains such as sketch and painting using the unified classifier. However, in other domains like clipart, quickgraph, and real, the accuracy was lower than when models were trained independently for each domain.}
The results in the fourth row show that by changing just the task-specific batchnorm layers in the jointly trained backbone, we can achieve better results than TAPS and Adashare. Our FTN with just $R=1$ also outperforms Adashare and TAPS. \

\section{Analysis on $\Delta W_t$ in FTN vs post-training factorization}
We have done experimental results to show that FTN learns explicit low-rank representation instead of a module that stores low-rank tensors. For post-training factorization, we consider  $\Delta W_t = W_t - W$, i.e. computing the difference between the weights of the fine-tuned network ($W_t$) and the original network ($W$). We have compressed or performed low-rank approximation on the trained $\Delta W_t$ using CP, Tensor train (TT), and SVD matrix factorization. We performed experiments on ImageNet-to-Sketch dataset with resnet-50 and efficient-net backbone with $R\in\{1,10,20,30,40,50\}$. Table~\ref{tab:deltaw_imagnet_to_sketch_resnet50} depicts that low-rank approximation of trained $\Delta W_t$ causes severe performance degradation. For instance, R=1, average accuracy for ImageNet-to-Sketch with resnet-50 drops to nearly 27\%. As R increases, the accuracy improves for all the factorization approaches, but they remain lower than FTN results. From Table~\ref{tab:deltaw_imagnet_to_sketch_efficientnet} the difference between FTN and post-training factorization is even more striking in the case of Efficient-Net. For instance, R=1, average accuracy drops to nearly 5\%. As R increases, the number of parameters increases sharply but even with 2x more parameters, the average accuracy of post-training low-rank factorization remains lower than FTN with R=1. The Table~\ref{tab:error_deltaw_imagnet_to_sketch_resnet50}, \ref{tab:error_deltaw_imagnet_to_sketch_efficientnet} depicts the approximation error (averaged over all the layers) for different values R and every task. We report average error and standard deviation over the layers for a subset of R. We observe that as R increases approximation error reduces, which leads to improvement in accuracy. Nevertheless, approximation error remains significant even with large R, which is the reason why accuracy remains significantly less than fine-tuned weights. In all cases, accuracy for tasks depends on approximation error (Flowers and CUBS do better than Wikiart and Sketch because the approximation error is small).
 
\begin{table*}[htpb]
    \centering
    \small
    \caption{Low-rank factorization of weight increments on ImageNet-to-Sketch dataset with resnet-50 backbone.}
    \label{tab:deltaw_imagnet_to_sketch_resnet50}
    \resizebox{0.85\linewidth}{!}{%
    \begin{tabular}{llllllll}
    \toprule
    \textbf{Method}                        & \textbf{Params (Abs)} & \textbf{Flowers} & \textbf{Wikiart} & \textbf{Sketch} & \textbf{Cars} & \textbf{CUB} & \textbf{mean} \\
    \midrule
    Fine-Tuning(arbitrary $\Delta W_t$ = $W_t$ -$W$) & 6x(141M)           & 95.69            & 78.42            & 81.02            & 91.44         & 83.32        & 85.98         \\
    FTN, R=1                               & 1.004x(23.95M)        & 94.79            & 73.03            & 78.62           & 86.85         & 80.86        & 82.83         \\
    FTN, R=50                              & 1.53x(36.02M)         & 96.42            & 78.01            & 80.6            & 90.83         & 82.96        & 85.76         \\
    \midrule
    $\Delta W$ with R=1 (CP)                       & 1.009x(23.71M)        & 64.73            & 5.36             & 6.6             & 10.17         & 51.97        & 27.76         \\
    $\Delta W$ with R=10 (CP)                      & 1.09x(25.67M)         & 88.58            & 10.97            & 48.02           & 56.42         & 70.59        & 54.91         \\
    $\Delta W$ with R=20 (CP)                      & 1.185x(27.84M)        & 92.76            & 20.83            & 61.88           & 78.65         & 76.98        & 66.22         \\
    $\Delta W$ with R=30 (CP)                      & 1.27x(30.02M)         & 94.15            & 37.68            & 71.27           & 83.56         & 79.81        & 73.29         \\
    $\Delta W$ with R=40 (CP)                      & 1.37x(32.197)         & 94.6             & 46.33            & 74.73           & 86.74         & 81.2         & 76.72         \\
    $\Delta W$ with R=50 (CP)                      & 1.46x(34.37M)         & 94.75            & 56.65            & 77.32           & 87.95         & 82.03        & 79.74         \\
    $\Delta W$ with R=1 (TT)                       & 1.009x(23.71M)        & 64.55            & 5.38             & 7.38            & 10.32         & 51.78        & 27.88         \\
    $\Delta W$ with R=10 (TT)                      & 1.15x(27.18M)         & 88.4             & 8.87             & 52.33           & 57.64         & 70.14        & 55.47         \\
    $\Delta W$ with R=20 (TT)                      & 1.31x(30.86M)         & 92.73            & 22.26            & 62.7            & 78.34         & 77.01        & 66.60         \\
    $\Delta W$ with R=30 (TT)                      & 1.46x(34.53M)         & 94.13            & 37.66            & 70.9            & 84.16         & 80.22        & 73.41         \\
    $\Delta W$ with R=40 (TT)                      & 1.62x(38.21M)         & 94.68            & 50.13            & 75.05           & 87.12         & 81.34        & 77.66         \\
    $\Delta W$ with R=50 (TT)                      & 1.78x(41.89M)         & 94.89            & 57.24            & 77.55           & 88.17         & 82.24        & 80.01         \\
    $\Delta W$ with R=1 (SVD)                      & 1.01x(23.79M)         & 62.69            & 6.19             & 5.88            & 12.21         & 50.21        & 27.43         \\
    $\Delta W$ with R=10 (SVD)                     & 1.12x(26.42M)         & 88.75            & 12.3             & 52.00           & 54.21         & 71.26        & 55.70         \\
    $\Delta W$ with R=20 (SVD)                     & 1.24x(29.34M)         & 92.81            & 20.89            & 63.35           & 77.63         & 77.03        & 66.34         \\
    $\Delta W$ with R=30 (SVD)                     & 1.37x(32.26M)         & 94.18            & 35.2             & 69.77           & 83.65         & 79.48        & 72.45         \\
    $\Delta W$ with R=40 (SVD)                     & 1.49x(35.18M)         & 94.55            & 51.04            & 74.38           & 86.64         & 80.96        & 77.51         \\
    $\Delta W$ with R=50 (SVD)                     & 1.62x(38.13M)         & 94.89            & 58.92            & 76.45           & 87.82         & 81.71        & 79.95   \\
    \bottomrule
    \end{tabular}}
\end{table*}

\begin{table*}[htbp]
    \centering
    \small
    \caption{Average (std deviation) approximation error for low-rank factorization of weight increments $\Delta W$ on ImageNet-to-Sketch dataset with resnet-50 backbone.}
    \label{tab:error_deltaw_imagnet_to_sketch_resnet50}
    \resizebox{0.85\linewidth}{!}{%
    \begin{tabular}{lllllll}
    \toprule
    \textbf{Method}    & \textbf{Flowers} & \textbf{Wikiart} & \textbf{Sketch} & \textbf{Cars} & \textbf{CUB} & \textbf{mean} \\
    \midrule
    $\Delta W$ norm            & 0.88 (1.18)      & 31.15 (34.21)    & 10.89 (11.95)   & 2.23 (2.97)   & 1.66 (2.54)  & 9.36 (10.57)  \\
    \midrule
    $\Delta W$ with R=1 (CP)   & 0.85 (1.15)      & 30.56 (34.18)    & 10.64 (11.93)   & 2.16 (2.92)   & 1.60 (2.50)  & 9.16 (10.53)  \\
    $\Delta W$ with R=10 (CP)  & 0.63 (0.87)      & 26.53 (31.00)    & 9.07 (10.71)    & 1.72 (2.33)   & 1.27 (2.00)  & 7.84 (9.38)   \\
    $\Delta W$ with R=20 (CP)  & 0.49 (0.66)      & 23.14 (28.01)    & 7.81 (9.55)     & 1.40 (1.87)   & 1.03 (1.58)  & 6.77 (8.33)   \\
    $\Delta W$ with R=30 (CP)  & 0.40 (0.52)      & 20.38 (25.42)    & 6.81 (8.55)     & 1.17 (1.85)   & 0.85 (1.29)  & 5.922 (7.52)  \\
    $\Delta W$ with R=40 (CP)  & 0.33 (0.43)      & 18.11 (23.19)    & 6.00 (7.72)     & 1.003 (1.32)  & 0.73 (1.08)  & 5.23 (6.74)   \\
    $\Delta W$ with R=50 (CP)  & 0.27 (0.36)      & 16.20 (21.29)    & 5.34 (7.02)     & 0.87 (1.16)   & 0.63 (0.93)  & 4.66 (6.15)   \\
    $\Delta W$ with R=1 (TT)   & 0.85 (1.15)      & 30.59 (34.20)    & 10.66 (11.94)   & 2.16 (2.92)   & 1.60 (2.50)  & 9.17 (10.54)  \\
    $\Delta W$ with R=10 (TT)  & 0.62 (0.87)      & 25.90 (30.72)    & 8.84 (10.57)    & 1.68 (2.31)   & 1.24 (1.99)  & 7.65 (9.29)   \\
    $\Delta W$ with R=20 (TT)  & 0.48 (0.66)      & 22.11 (27.42)    & 7.45 (9.28)     & 1.33 (1.83)   & 0.98 (1.56)  & 6.47 (8.15)   \\
    $\Delta W$ with R=30 (TT)  & 0.38 (0.52)      & 19.08 (24.53)    & 6.37 (8.19)     & 1.09 (1.49)   & 0.80 (1.26)  & 5.54 (7.19)   \\
    $\Delta W$ with R=40 (TT)  & 0.30 (0.42)      & 16.58 (22.00)    & 5.50 (7.26)     & 0.91 (1.25)   & 0.66 (1.04)  & 4.79 (6.39)   \\
    $\Delta W$ with R=50 (TT)  & 0.25 (0.34)      & 14.50 (19.81)    & 4.78 (6.47)     & 0.77 (1.07)   & 0.56 (0.88)  & 4.17 (5.71)   \\
    $\Delta W$ with R=1 (SVD)  & 0.85 (1.15)      & 30.57 (34.21)    & 10.65 (11.93)   & 2.16 (2.92)   & 1.61 (2.50)  & 9.16 (10.54)  \\
    $\Delta W$ with R=10 (SVD) & 0.65 (0.88)      & 26.55 (31.23)    & 9.11 (10.78)    & 1.74 (2.36)   & 1.29 (2.02)  & 7.86 (9.45)   \\
    $\Delta W$ with R=20 (SVD) & 0.51 (0.69)      & 23.13 (28.36)    & 7.86 (9.67)     & 1.43 (1.92)   & 1.05 (1.61)  & 6.79 (8.45)   \\
    $\Delta W$ with R=30 (SVD) & 0.42 (0.56)      & 20.34 (25.84)    & 6.87 (8.72)     & 1.20 (1.61)   & 0.88 (1.33)  & 5.94 (7.61)   \\
    $\Delta W$ with R=40 (SVD) & 0.34 (0.46)      & 18.01 (23.63)    & 6.06 (7.90)     & 1.03 (1.38)   & 0.75 (1.13)  & 5.23 (6.9)    \\
    $\Delta W$ with R=50 (SVD) & 0.29 (0.39)      & 16.05 (21.69)    & 5.38 (7.21)     & 0.89 (1.22)   & 0.65 (0.98)  & 4.65 (6.29) \\
    \bottomrule
    \end{tabular}}
\end{table*}

\begin{table*}[htbp]
    \centering
    \small
    \caption{Low-rank factorization of weight increments on ImageNet-to-Sketch dataset with efficient-net backbone.}
    \label{tab:deltaw_imagnet_to_sketch_efficientnet}
    \resizebox{0.85\linewidth}{!}{%
    \begin{tabular}{llllllll}
    \toprule
    \textbf{Method}                      & \textbf{Params (Abs)} & \textbf{Flowers} & \textbf{Wikiart} & \textbf{Sketch} & \textbf{Cars} & \textbf{CUB} & \textbf{mean} \\
    \midrule
    Fine-Tuning (arbitrary $\Delta W_t$ = $W_t$ -$W$) & 6x(105.24M)           & 96.08            & 78.72            & 80.9            & 92.81         & 83.67        & 86.43         \\
    FTN, R=1                             & 1.079x(18.93M)        & 93.28            & 74.07            & 79.77           & 87.93         & 82.82        & 83.57         \\
    FTN, R=10                            & 1.474x(25.88M)        & 94.79            & 77.54            & 80.7            & 89.71         & 84.70        & 85.48         \\
    \midrule
    $\Delta W$ with R=1 (CP)                     & 1.04x(18.31M)         & 20.08            & 0.65             & 0.4             & 0.83          & 3.02         & 4.99          \\
    $\Delta W$ with R=10 (CP)                    & 1.43x(25.24M)         & 76.83            & 0.65             & 10.5            & 21.68         & 49.43        & 31.81         \\
    $\Delta W$ with R=20 (CP)                    & 1.87x(32.95M)         & 91.72            & 0.65             & 46.63           & 72.73         & 73.23        & 56.99         \\
    $\Delta W$ with R=30 (CP)                    & 2.31x(40.66M)         & 94.68            & 2.32             & 68.53           & 85.59         & 79.27        & 65.07         \\
    $\Delta W$ with R=40 (CP)                    & 2.75x(48.37M)         & 95.35            & 26.45            & 63.52           & 89.65         & 81.79        & 71.35         \\
    $\Delta W$ with R=50 (CP)                    & 2.79x(49.02M)         & 95.45            & 41.42            & 76.5            & 91.03         & 82.34        & 77.34         \\
    $\Delta W$ with R=1 (TT)                     & 1.04x(18.31M)         & 20.00            & 0.65             & 0.4             & 0.8           & 3.52         & 5.07          \\
    $\Delta W$ with R=10 (TT)                    & 1.43x(25.2M)          & 77.02            & 0.65             & 11.17           & 22.62         & 49.19        & 32.13         \\
    $\Delta W$ with R=20 (TT)                    & 1.84x(32.3M)          & 91.49            & 0.66             & 47.5            & 72.94         & 72.51        & 57.02         \\
    $\Delta W$ with R=30 (TT)                    & 2.20x(38.76M)         & 94.78            & 2.37             & 68.38           & 86.08         & 79.46        & 66.21         \\
    $\Delta W$ with R=40 (TT)                    & 2.52x(44.3M)          & 95.32            & 31.64            & 74.6            & 89.79         & 81.95        & 74.66         \\
    $\Delta W$ with R=50 (TT)                    & 2.80x(49.22M)         & 95.63            & 63.18            & 78.5            & 91.07         & 82.84        & 82.24         \\
    $\Delta W$ with R=1 (SVD)                    & 1.07x(18.77M)         & 23.74            & 0.65             & 0.4             & 1.02          & 6.02         & 6.366         \\
    $\Delta W$ with R=10 (SVD)                   & 1.51x(26.59M)         & 76.84            & 0.65             & 11.33           & 22.81         & 49.43        & 32.21         \\
    $\Delta W$ with R=20 (SVD)                   & 1.86x(32.68M)         & 91.59            & 0.66             & 47.62           & 72.96         & 72.52        & 57.07         \\
    $\Delta W$ with R=30 (SVD)                   & 2.20x(38.63M)         & 94.78            & 2.37             & 68.38           & 86.08         & 79.46        & 66.21         \\
    $\Delta W$ with R=40 (SVD)                   & 2.51x(44.14M)         & 95.32            & 31.64            & 74.6            & 89.75         & 81.95        & 74.65         \\
    $\Delta W$ with R=50 (SVD)                   & 2.79x(49.09M)         & 95.63            & 63.18            & 78.5            & 91.07         & 82.84        & 82.24      \\
    \bottomrule
    \end{tabular}}
\end{table*}

\begin{table*}[!h]
    \centering
    \small
    \caption{Average (std deviation) approximation error for low-rank factorization of weight increments $\Delta W$ on ImageNet-to-Sketch dataset with efficient-net backbone.}
    \label{tab:error_deltaw_imagnet_to_sketch_efficientnet}
    \resizebox{0.85\linewidth}{!}{%
    \begin{tabular}{lllllll}
    \toprule
    \textbf{Method}    & \textbf{Flowers} & \textbf{Wikiart} & \textbf{Sketch} & \textbf{Cars} & \textbf{CUB} & \textbf{mean} \\
    \midrule
    $\Delta W$ norm            & 0.88 (1.18)      & 31.15 (34.21)    & 10.89 (11.95)   & 2.23 (2.97)   & 1.66 (2.54)  & 9.36 (10.57)  \\
    \midrule
    $\Delta W$ with R=1 (CP)   & 0.85 (1.15)      & 30.56 (34.18)    & 10.64 (11.93)   & 2.16 (2.92)   & 1.60 (2.50)  & 9.16 (10.53)  \\
    $\Delta W$ with R=10 (CP)  & 0.63 (0.87)      & 26.53 (31.00)    & 9.07 (10.71)    & 1.72 (2.33)   & 1.27 (2.00)  & 7.84 (9.38)   \\
    $\Delta W$ with R=20 (CP)  & 0.49 (0.66)      & 23.14 (28.01)    & 7.81 (9.55)     & 1.40 (1.87)   & 1.03 (1.58)  & 6.77 (8.33)   \\
    $\Delta W$ with R=30 (CP)  & 0.40 (0.52)      & 20.38 (25.42)    & 6.81 (8.55)     & 1.17 (1.85)   & 0.85 (1.29)  & 5.922 (7.52)  \\
    $\Delta W$ with R=40 (CP)  & 0.33 (0.43)      & 18.11 (23.19)    & 6.00 (7.72)     & 1.003 (1.32)  & 0.73 (1.08)  & 5.23 (6.74)   \\
    $\Delta W$ with R=50 (CP)  & 0.27 (0.36)      & 16.20 (21.29)    & 5.34 (7.02)     & 0.87 (1.16)   & 0.63 (0.93)  & 4.66 (6.15)   \\
    $\Delta W$ with R=1 (TT)   & 0.85 (1.15)      & 30.59 (34.20)    & 10.66 (11.94)   & 2.16 (2.92)   & 1.60 (2.50)  & 9.17 (10.54)  \\
    $\Delta W$ with R=10 (TT)  & 0.62 (0.87)      & 25.90 (30.72)    & 8.84 (10.57)    & 1.68 (2.31)   & 1.24 (1.99)  & 7.65 (9.29)   \\
    $\Delta W$ with R=20 (TT)  & 0.48 (0.66)      & 22.11 (27.42)    & 7.45 (9.28)     & 1.33 (1.83)   & 0.98 (1.56)  & 6.47 (8.15)   \\
    $\Delta W$ with R=30 (TT)  & 0.38 (0.52)      & 19.08 (24.53)    & 6.37 (8.19)     & 1.09 (1.49)   & 0.80 (1.26)  & 5.54 (7.19)   \\
    $\Delta W$ with R=40 (TT)  & 0.30 (0.42)      & 16.58 (22.00)    & 5.50 (7.26)     & 0.91 (1.25)   & 0.66 (1.04)  & 4.79 (6.39)   \\
    $\Delta W$ with R=50 (TT)  & 0.25 (0.34)      & 14.50 (19.81)    & 4.78 (6.47)     & 0.77 (1.07)   & 0.56 (0.88)  & 4.17 (5.71)   \\
    $\Delta W$ with R=1 (SVD)  & 0.85 (1.15)      & 30.57 (34.21)    & 10.65 (11.93)   & 2.16 (2.92)   & 1.61 (2.50)  & 9.16 (10.54)  \\
    $\Delta W$ with R=10 (SVD) & 0.65 (0.88)      & 26.55 (31.23)    & 9.11 (10.78)    & 1.74 (2.36)   & 1.29 (2.02)  & 7.86 (9.45)   \\
    $\Delta W$ with R=20 (SVD) & 0.51 (0.69)      & 23.13 (28.36)    & 7.86 (9.67)     & 1.43 (1.92)   & 1.05 (1.61)  & 6.79 (8.45)   \\
    $\Delta W$ with R=30 (SVD) & 0.42 (0.56)      & 20.34 (25.84)    & 6.87 (8.72)     & 1.20 (1.61)   & 0.88 (1.33)  & 5.94 (7.61)   \\
    $\Delta W$ with R=40 (SVD) & 0.34 (0.46)      & 18.01 (23.63)    & 6.06 (7.90)     & 1.03 (1.38)   & 0.75 (1.13)  & 5.23 (6.9)    \\
    $\Delta W$ with R=50 (SVD) & 0.29 (0.39)      & 16.05 (21.69)    & 5.38 (7.21)     & 0.89 (1.22)   & 0.65 (0.98)  & 4.65 (6.29)  \\
    \bottomrule
    \end{tabular}}
    
\end{table*}

\section{Multi-domain image generation}

A deep generative network $\mathbf{G}$, parameterized by $\mathbf{\mathcal{W}}$, can learn to map a low-dimensional latent code $\mathbf{z}$ to a high-dimensional natural image $\mathbf{x}$ \cite{ulyanov2018deep,zhu2016generative, pan2021exploiting}. To find a latent representation for a set of images given a pre-trained generative network, we can solve the following optimization problem:
\begin{equation}\label{eq:basic_img_gen}
    \min_{z_{i}} \sum_{i=1}^{N}  \left\| x_i - \mathbf{G}(z_i; \mathcal{W}) \right\|_p^p. 
\end{equation}
The work in~\cite{pan2021exploiting} as well as our experimental results show that this approach is very limited in handling complex and diverse images. If $\mathbf{x}$ is an image that belongs to a domain that is different from the source domain used to train the generator, we are not guaranteed to find a latent vector $\mathbf{z}^*$ such that $\mathbf{x} \approx \mathbf{G}(\mathbf{z}^*)$. 
We showed FTNs can be used to expand the range of $\mathbf{G}$ by reparametrizing it as $\mathbf{G}(z_i; \mathcal{W}, \Delta \mathcal{W}_t)$, where $\Delta \mathcal{W}_t$ are the domain-specific low-rank factors and Batch Normalization parameters.  By optimizing over the latent vectors $z_i$ and domain specific parameters, we learned to generate images from new domains. \\

\noindent \textbf{Dataset.} We used the multi-domain Transient Attributes \cite{Laffont14} dataset. The dataset contains a total of 8571 images with 40 annotated attribute labels. Each label is associated with a score in the $[-1,1]$ range. We utilize the associated confidence score for each season to build our collection of images for each season. Additionally, we only selected images that were captured during daytime. Our training set consisted of $1875$ summer, $1405$ spring, $1353$ autumn, and $2566$ winter images. We normalized each image to the range of $[-1,1]$ and resized them to a resolution of $128\times128$. \\

\noindent \textbf{Training details.} Our base network follows the BigGAN architecture \cite{brocklarge2019} that was pre-trained for 100k iterations on ImageNet using $128\times128$ images. We trained all the networks in this experiment using Adam optimizer. We used a learning rate of $0.05$ for the low-rank tensors and a learning rate of $0.001$ for the linear layers. We did not use any weight decay. We trained for $2000$ epochs with a cosine annealing learning rate scheduler and an early stopping criterion ranging from $200$ to $600$ iterations. \\

\begin{figure*}[htbp]
  \centering
  \includegraphics[width=0.5\linewidth]{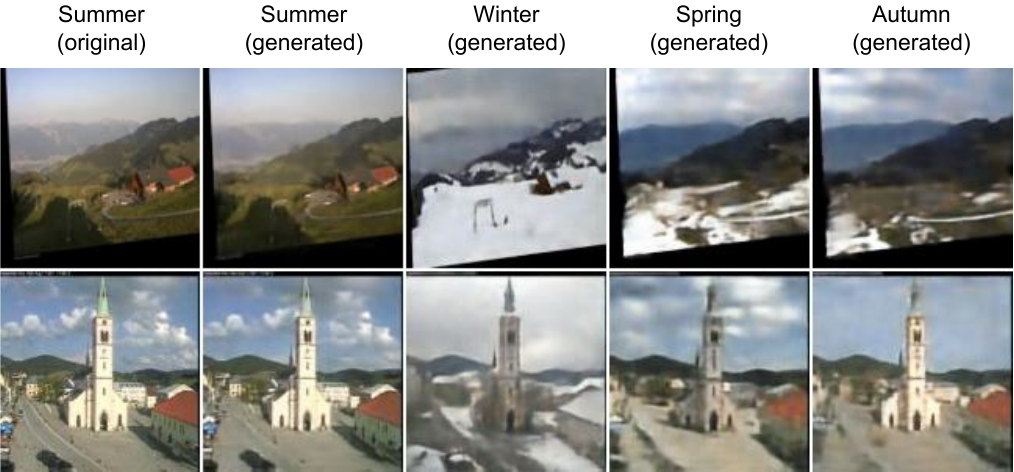}
  \captionof{figure}{Generated images for different seasons using FTN. }
  \label{fig:seasons_collage}
\end{figure*}

\begin{figure*}[ht]
  \centering
  \includegraphics[width=0.55\linewidth]{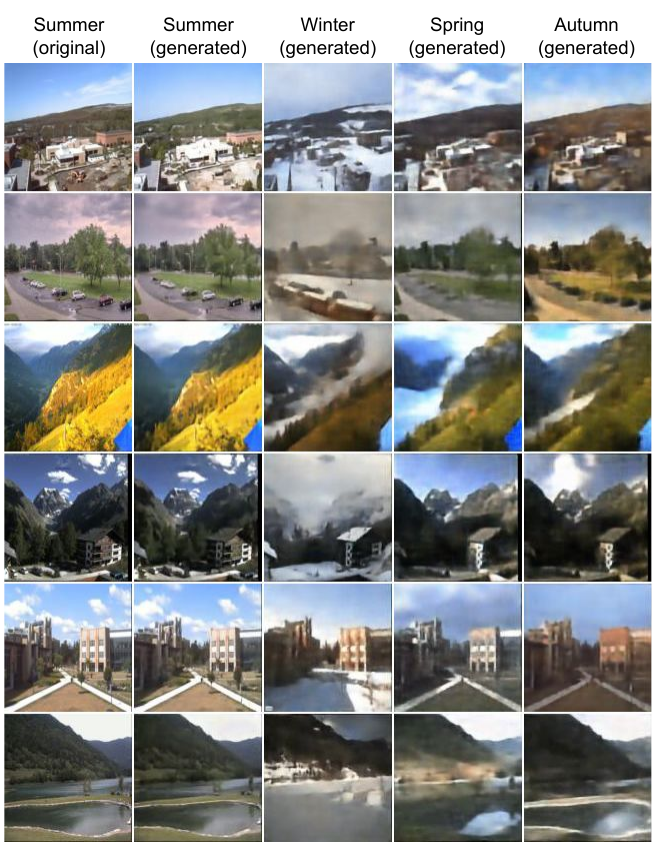}
  \captionof{figure}{Additional generated images for different seasons using FTN. }
  \label{fig:additional_seasons_collage}
  \vspace{-2em}
\end{figure*}

\noindent \textbf{Results.} Table \ref{tab:img_gen_dom_adapt} shows the performance, in terms of Peak Signal-to-Noise Ratio (PSNR), for three methods. Our proposed FTN network achieved a comparable performance to the single-domain networks. Each single domain network has $71.4$M trainable parameters, while the FTN network adds an additional $3.9$M parameters per domain over the base network. Figure~\ref{fig:seasons_collage} and \ref{fig:additional_seasons_collage} show examples of images generated by our proposed FTN network. In addition, table \ref{tab:diff_rank_img_gen_dom_adapt} shows the average performance of our FTN network under different rank settings. We observe a performance increase by increasing the rank $R$ of our low-rank factors.

\begin{table}[h]
    \centering
    \caption{Image generation PSNR for different methods and seasons}
    \label{tab:img_gen_dom_adapt}
    \begin{tabular}{lccc}
        \toprule
        \textbf{Season} & \textbf{Single domain}  & \textbf{FTN}   \\
        \midrule
        Summer  & 25.30           & 25.30  \\
        Winter  & 21.30           & 22.23 \\
        Spring   & 22.07           & 20.50  \\
        Autumn   & 19.87           & 20.08 \\
        \bottomrule
    \end{tabular}
\end{table}
\begin{table}[h]
    \centering
    \caption{FTN PSNR for image generation under $R=\{20, 50\}$}
    \label{tab:diff_rank_img_gen_dom_adapt}
    \begin{tabular}{lccc}
        \toprule
        \textbf{Season} &  \textbf{Rank 20} & \textbf{Rank 50}  \\
        \midrule
        Winter &  18.11  & 22.23 \\
        Spring &  18.89 & 20.50  \\
        Autumn &  17.95 & 20.08 \\
        \bottomrule
    \end{tabular}
    \vspace{-2em}
\end{table}

